\documentclass[10pt,journal,compsoc]{IEEEtran}
\ifCLASSOPTIONcompsoc
  \usepackage[nocompress]{cite}
\else
  \usepackage{cite}
\fi

\ifCLASSINFOpdf
\else
\fi

\usepackage[utf8]{inputenc} 
\usepackage[T1]{fontenc}    
\usepackage{hyperref}       
\usepackage{url}            
\usepackage{booktabs}       
\usepackage{amsfonts}       
\usepackage{nicefrac}       
\usepackage{microtype}      
\usepackage{xcolor}         
\usepackage{amsmath}
\usepackage{multirow}
\usepackage{float}
\usepackage{colortbl}
\usepackage{graphicx} 
\usepackage{subcaption}

\usepackage{graphicx}
\usepackage{arydshln}
\usepackage{caption}
\usepackage{tabularx}
\usepackage{adjustbox}
\usepackage{algorithmic}
\usepackage{xcolor}
\definecolor{dark green}{rgb}{0.0, 0.5, 0.0}
\usepackage[linesnumbered,boxed,ruled,commentsnumbered]{algorithm2e}

\begin{document}
\renewcommand\arraystretch{1.5}
%
\title{\title{Robust Disentangled Counterfactual Learning for Physical Audiovisual Commonsense Reasoning}}
%
%
%
%

\author{
        Mengshi Qi,~\IEEEmembership{Member,~IEEE},
        Changsheng Lv,
        Huadong Ma,~\IEEEmembership{Fellow,~IEEE}
\thanks{This work is partly supported by the Funds for the NSFC Project under Grant 62202063, Beijing Natural Science Foundation (L243027). (\emph{Corresponding author: Mengshi Qi~(email:~qms@bupt.edu.cn)})}
\thanks{M. Qi, C. Lv, and H. Ma are with the State Key Laboratory of Networking and Switching Technology, Beijing University of Posts and Telecommunications, China.}
}
%
%

\markboth{}%
{Shell \MakeLowercase{\textit{et al.}}: Bare Demo of IEEEtran.cls for Computer Society Journals}
%



\IEEEtitleabstractindextext{%
\begin{abstract}
In this paper, we propose a new Robust Disentangled Counterfactual Learning~(RDCL) approach for physical audiovisual commonsense reasoning. The task aims to infer objects' physics commonsense based on both video and audio input, with the main challenge being how to imitate the reasoning ability of humans, even under the scenario of missing modalities. Most of the current methods fail to take full advantage of different characteristics in multi-modal data, and lacking causal reasoning ability in models impedes the progress of implicit physical knowledge inferring. 
To address these issues, our proposed RDCL method decouples videos into static (time-invariant) and dynamic (time-varying) factors in the latent space by the disentangled sequential encoder, which adopts a variational autoencoder (VAE) to maximize the mutual information with a contrastive loss function. Furthermore, we introduce a counterfactual learning module to augment the model's reasoning ability by modeling physical knowledge relationships among different objects under counterfactual intervention. To alleviate the incomplete modality data issue, we introduce a robust multimodal learning method to recover the missing data by decomposing the shared features and model-specific features. Our proposed method is a plug-and-play module that can be incorporated into any baseline including VLMs. In experiments, we show that our proposed method improves the reasoning accuracy and robustness of baseline methods and achieves the state-of-the-art performance. Our code and data are available at https://github.com/MICLAB-BUPT/DCL.
\end{abstract}

}

\maketitle

\IEEEdisplaynontitleabstractindextext

%
\IEEEpeerreviewmaketitle

\IEEEraisesectionheading{\section{Introduction}\label{sec:introduction}}

%
%
%
%

\begin{figure}[!t]
    \centering
    \includegraphics[width=0.9\linewidth]{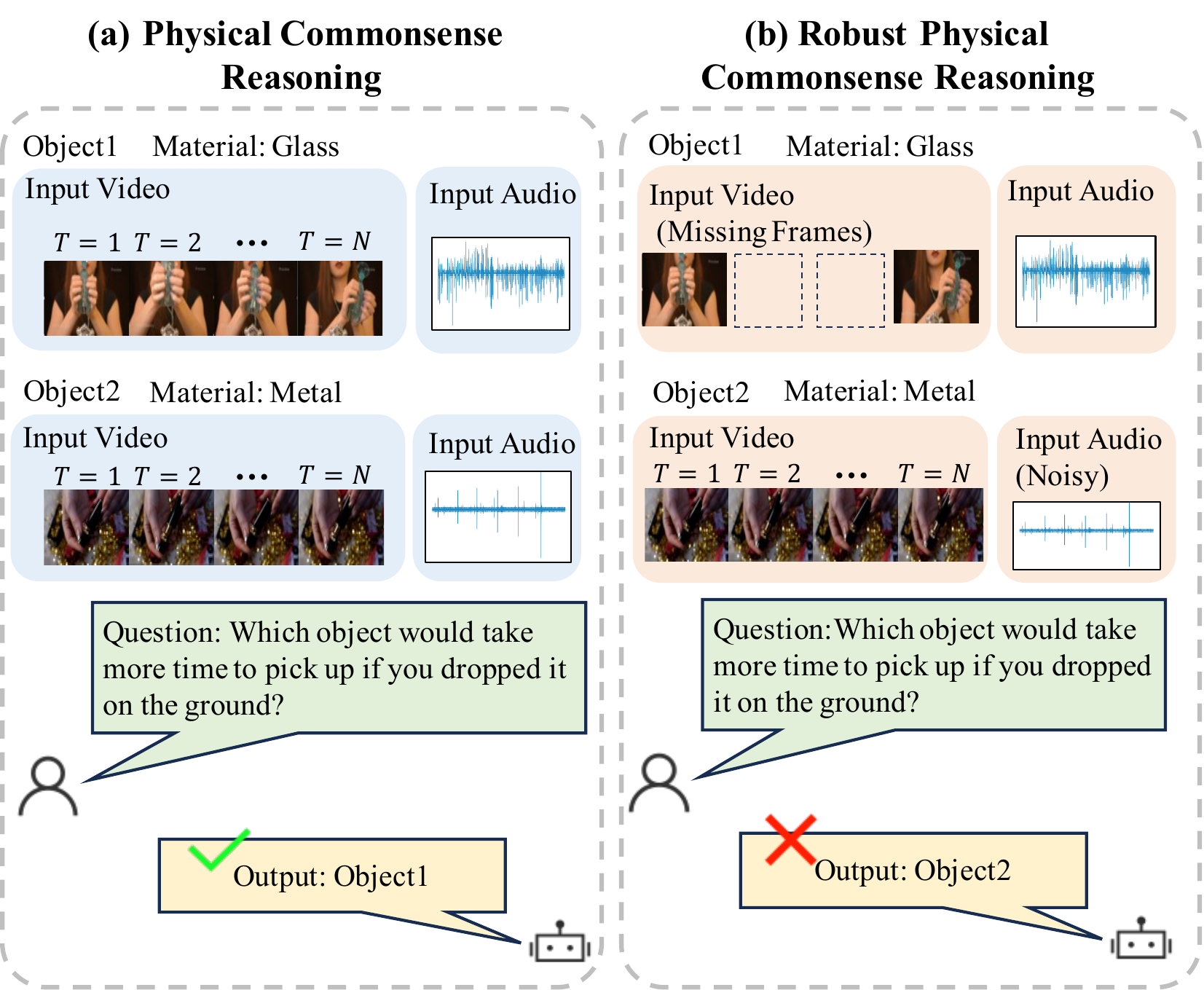}
    \vspace{-2mm}
    \caption{Illustration of our main tasks. Task (a) involves AVQA for physical commonsense reasoning, while task (b) addresses robust AVQA, which deals with missing modality data for physical commonsense reasoning encountered in real-world scenarios.}
    \label{fig1:Illustration}
    \vspace{-4mm}
\end{figure}

\IEEEPARstart{H}{umans} acquire the physical commonsense knowledge by integrating information from various modalities, enabling them to deduce the properties of unfamiliar objects in the daily life~\cite{kriegeskorte2015deep}. This includes tasks such as determining material composition (e.g., "this object is likely made of wood") or solving practical problems (e.g., "which object would cause a greater mess if it fell") \cite{yu2022pacs}. Such reasoning remains a significant challenge for machine intelligence, yet it is essential for applications like robot navigation \cite{an2024etpnav} and augmented or virtual reality systems. In this paper, we employ Audio-Visual Question Answering~(AVQA) as a proxy task to advance the machine’s capacity for physical commonsense reasoning.
As shown in Figure~\ref{fig1:Illustration}(a), the AVQA aimed to select the correct answer to the question by comparing the given two objects. For each object, our input consists of a video of human-object interactions and its corresponding audio.

The major challenge in audio-visual physical commonsense reasoning lies in effectively extracting and reasoning about implicit physical knowledge from the vast amounts of multi-modal data, particularly from videos. This necessitates the capability to analyze intricate video content, recognize the categories and associated physical attributes of various objects, and comprehend the causal interactions between them. These cognitive functions closely resemble how humans acquire knowledge, learn, and reason about the physical environment.

Current existing methods~\cite{purushwalkam2021audio,chen2021semantic} typically extract generic visual features from videos depicting human-object interactions, resulting in mixed feature representations that fail to separate object and action information. This approach often results in misidentifying relevant objects due to insufficient contextual detail. However, in physical commonsense reasoning, it is crucial to identify the attributes and physical properties of objects. 

To address this challenge, we propose an approach to disentangle video content into two distinct factors: static factors, which remain constant over time, and dynamic factors, which change over time. Another motivation for our paper is to establish relationships of physical knowledge among different objects across both video and audio modalities. We improve the optimization of our results by considering the relevance of multiple samples and integrating causal learning, using these relationships as confounders. Additionally, the implementation of counterfactual interventions enhances the model's explainability and reasoning capabilities.

Furthermore, current methods infer the physical commonsense under the assumption of modality completeness. However, a few real-world factors such as particular modality data missing~\cite{hou2019deep,peng2022balanced} invariably bring robustness challenges, As shown in Figure~\ref{fig1:Illustration}(b). For instance, privacy restrictions in mobile applications or sensor corruptions in robot navigation can result in data limitations or low-quality data, respectively. To ensure robust physical commonsense reasoning in such scenarios, we further extend the work to learn the relationship between audio and video representations, and then recover missing modal information through shared features across these modalities.

In this paper, we propose a novel approach for audiovisual physical commonsense reasoning, named \emph{\textbf{D}isentangled} \emph{\textbf{C}ounterfactual} \emph{\textbf{L}earning} (DCL). It explicitly extracts static and dynamic factors from video and employs causal learning to reveal physical knowledge relationships among various objects. To achieve this, we design a Disentangled Sequential Encoder~(DSE), which utilizes a sequential variational autoencoder for effectively self-supervised separation of static and dynamic video factors. Additionally, we incorporate a contrastive estimation method to enhance the mutual information (MI) between the input data and the two latent factors, while simultaneously reducing MI between static and dynamic factors. Furthermore, we introduce a novel Counterfactual Learning Module~(CLM) to capture physical knowledge relationships from a diverse range of data samples by counterfactual interventions. The model's training objectives are refined by maximizing the probability likelihood in the DSE and the Total Indirect Effect value in the CLM.

More importantly, this paper extends our NeurIPS conference paper~\cite{lv2024disentangled}, enhancing the DCL to \emph{\textbf{R}obust} \emph{\textbf{D}isentangled} \emph{\textbf{C}ounterfactual} \emph{\textbf{L}earning}~(RDCL). In contrast to the original version, we have devised a novel method to improve DSE, by computing discriminative information between positive and negative samples. Moreover, to address the challenge of missing modality data in real-world scenarios, we incorporate a new incomplete multi-modal learning method, which extracts shared semantic information representing physical knowledge across modalities, and supplements missing modalities during both training and testing, by leveraging shared semantic features from other modalities. In our experiments, we evaluate both DCL and RDCL on the PACS dataset~\cite{yu2022pacs}, and further conduct comprehensive tests, present additional visualizations, and perform more detailed ablation studies to demonstrate the effectiveness of each proposed component in our approach. In addition, we analyze and discuss about the visual bias and VLM-assisted reasoning issues. 

Our main contributions can be summarized as follows:

\par\textbf{(1)} We introduce a novel Disentangled Counterfactual Learning (DCL) approach for physical audiovisual commonsense reasoning, which separates video inputs into static and dynamic factors using a Disentangled Sequential Encoder. 

\par\textbf{(2)} We present a new Counterfactual Learning Module designed to model physical knowledge relationships among various objects, utilizing these relationships as counterfactual interventions to enhance causal reasoning capabilities.

\par\textbf{(3)} We design a new Robust Disentangled Counterfactual Learning~(RDCL) method, which decomposes multimodal data into modality-shared information among various modalities data and modality-specific information and utilizes the shared information between modalities to complete missing modalities.  

\par\textbf{(4)} We conducted comprehensive comparisons with other methods on the PACS dataset under both complete and incomplete modality conditions. Compared to DCL, our RDCL achieves relative improvements of 3.3\% on the complete PACS dataset and 11.8\% on the PACS dataset under incomplete.

\section{Related Work}

{\bf Physical Commonsense Reasoning.}~Commonsense knowledge, embedded in a variety of data, is acquired by humans and used for reasoning about unseen things~\cite{piloto2022intuitive}. Hespos {\it et al.}~\cite{hespos2016five} show that infants' commonsense aids in reasoning about knowledge, and machines can similarly learn and perform well on physical knowledge~\cite{piloto2022intuitive}. Machines can acquire commonsense from various data types, including visual~\cite{zellers2019recognition,li2022representation}, textual~\cite{bisk2020piqa}, audio\cite{zellers2022merlot}, and multimodal data~\cite{yu2022pacs}. 
Zellers~{\it et al.}~\cite{zellers2019recognition} constructed a visual question-answering~(VQA) dataset for visual commonsense reasonng~(VCR), guiding models to utilize learned commonsense knowledge for high-level cognition and reasoning beyond images. Wang~{\it et al.}\cite{wang2020visual} proposed an unsupervised approach to mine visual commonsense, enhancing model performance on visual captioning and VQA. Zareian~{\it et al.} ~\cite{zareian2020learning} proposed the first method to automatically acquire visual commonsense such as affordance and intuitive physics from data for scene graph generation. Li~{\it et al.}~\cite{li2022representation} further introduced a video-based VQA dataset, \emph{Video-VQA}, which not only involves reasoning questions about video content but also generates appropriate justifications based on commonsense knowledge. Bisk~{\it et al.}~\cite{bisk2020piqa} firstly proposed the task of learning physical commonsense from text and constructed a corresponding benchmark dataset, \emph{PIQA}. Lin~{\it et al.}~\cite{lin2023tiktalk} explored the usage of commonsense knowledge in human-like chatbots with multi-modal context. However, most work focused on learning visual and audio commonsense knowledge, with a lack of learning the physics from visual objects. Yu~{\it et al.}~\cite{yu2022pacs} introduced a multimodal physical commonsense knowledge dataset based on visual, audio, and text, \emph{PACS}, and performed the VQA task related to the physical commonsense in a fusion manner. In contrast, our proposed method decouples physical commonsense into static and dynamic aspects and introduces causal learning to enhance reasoning ability for physical problems. 

\noindent
{\bf Disentangled Representation Learning~(DRL).}~DRL aims to learn various hidden explanatory factors behind observable data~\cite{bengio2013representation}, and it has been widely applied in computer vision~\cite{chen2021curriculum}, including image recognition~\cite{wei2024unsupervised,qi2020stc}, visual reasoning~\cite{van2019disentangled,qi2021semantics}, and generation~\cite{ma2018disentangled, bai2021contrastively, wang2024rdfc,qi2019attentive,qi2019ke},. Tran~{\it et al.}~\cite{tran2017disentangled} employed a Generative Adversarial Network (GAN)~\cite{goodfellow2020generative} to explicitly disentangle facial variations, addressing face recognition across diverse human poses.  Similarly, Wei~{\it et al.}\cite{wei2024unsupervised} utilized a Variational Autoencoder~(VAE) to disentangle actions within videos, enhancing unsupervised cross-domain action recognition by decoupling videos into domain-specific and domain-invariant features. Moreover, disentangled representation has been leveraged in image generation. Ma~{\it et al.} ~\cite{ma2018disentangled} disentangled images into foreground, background, and pose information, generating new person images based on these manipulated factors through a multi-branch reconstruction network and adversarial training. Differing from static image processing, Bai~{\it et al.}~\cite{bai2021contrastively} and Zhu~{\it et al.}~\cite{zhu2020s3vae} investigated video generation by disentangling and merging static and dynamic character information. Wang~{\it et al.}\cite{wang2023disavr} addressed the visual semantic ambiguity problem by decoupling questions into region-related, spatial-related, and semantic-related features. Contrary to previous methods that explicitly model disentangled factors, our work centers on learning the relationships of physical knowledge across different samples. We utilize this knowledge to assist in answering relevant questions, thereby enhancing the model's interpretability.

\noindent{\bf Causal learning.}~Due to the "language prior"~\cite{goyal2017making} or "visual  bias"~\cite{antol2015vqa} in traditional VQA datasets, current methods rely heavily on inherent biases in language or visual features, leading to inaccurate answers. Recently, counterfactual causal reasoning have been utilized in VQA~\cite{goyal2019counterfactual}, scene graph generation~\cite{tang2020unbiased}, image recognition~\cite{Rao_2021_ICCV}, and video understanding~\cite{xu2022unintentional}.  These techniques not only mitigate the impact of data biases on results~\cite{sun2023unbiased}, but also enhance model interpretability during inference~\cite{xue2023variational}. Different from the current work~\cite{li2023progressive} focusing VQA with cross-modal modeling, our approach distinctively concentrates on constructing physical knowledge relationships among different samples and employing them as confounders in causal reasoning.

\noindent{\bf Roubst multimodal learning.}~Multimodal learning encompasses various types of data, including visual-text~\cite{lu2019vilbert}, visual-audio~\cite{hu2021class}, text-audio~\cite{deshmukh2023pengi}, and visual-text-audio~\cite{li2022learning}. However in practical applications, data from different modalities may exhibit varying degrees of missing information~\cite{wang2023multi}, which can lead to the performance decrease of multimodal systems, sometimes even inferior to those of the single-modal approach. In this work, we propose a new robust multi-modal learning that aligns the shared semantic information across different modalities and then utilizes this information to complete the missing modality.

\begin{figure*}[!t]
    \centering
    \includegraphics[width=0.82\linewidth]{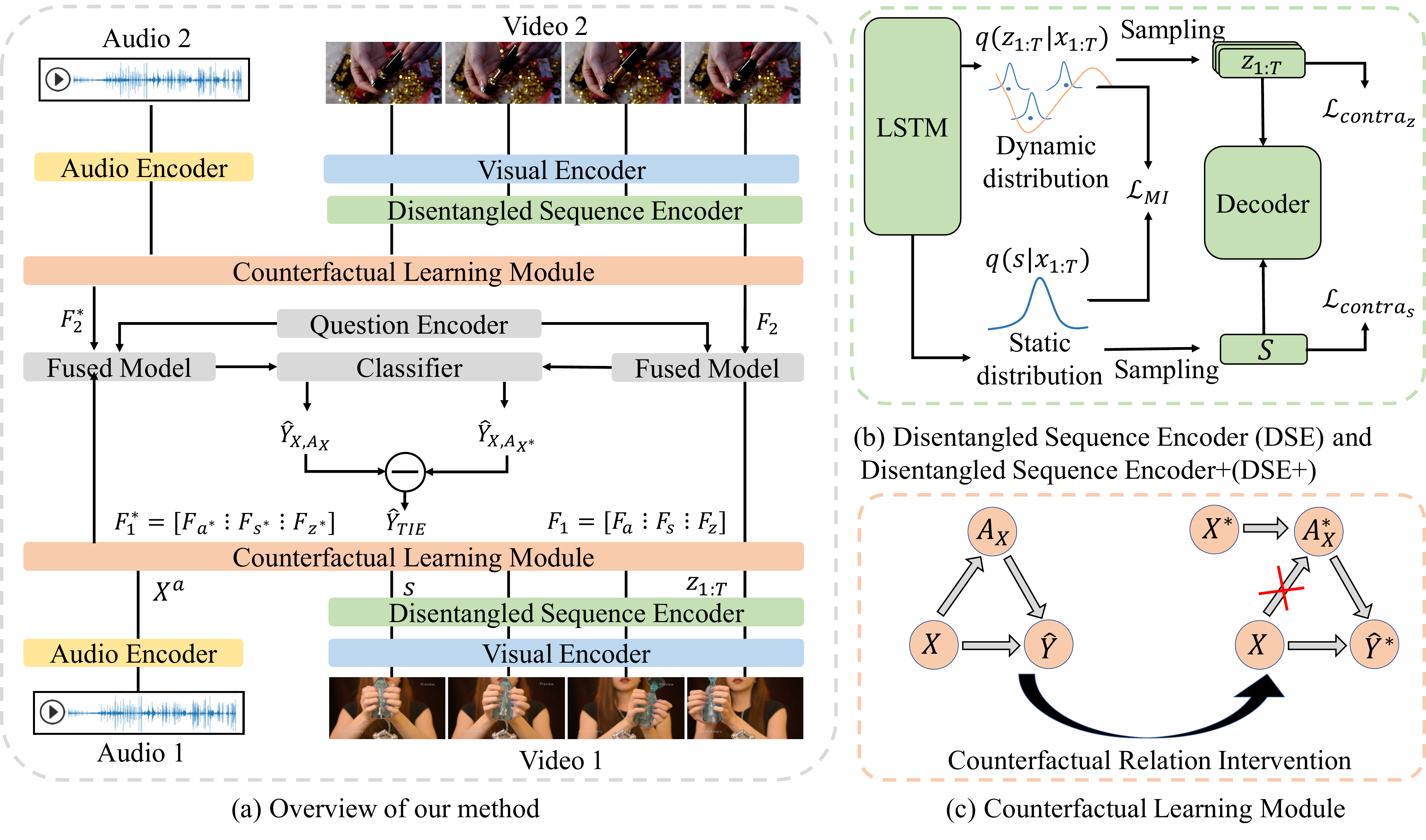}
    \vspace{-4mm}
    \caption{The illustration of our proposed DCL model: Part (a) presents the overall structure, which begins with the input of videos accompanied by audio. These are initially encoded via the respective visual and audio encoders. Subsequently, the Disentangled Sequence Encoder in Part (b) is employed to segregate video features into static and dynamic elements utilizing an LSTM-based Variational Autoencoder (VAE). The Counterfactual Learning Module in Part (c) is then used to construct the affinity matrix `A', which acts as a confounder, and to derive the prediction $\hat{Y}_{X, A_X}$ and the counterfactual outcome $\hat{Y}_{X, A^*_X}$. Ultimately, we compute $\hat{Y}_{TIE}$ by subtracting these two outcomes and optimizing the model.}
    \label{fig: an overview of model}
    \vspace{-3mm}
\end{figure*}

\section{Problem Defination}
\label{sec:problem}

The task of physical audiovisual commonsense reasoning involves executing a binary classification. It requires the model to extract features from the audio and video associated with two distinct objects, and subsequently select the most appropriate one in response to a specific question. A pair of videos, denoted as $<v_1, v_2>$, and their corresponding audios, denoted as $<a_1, a_2>$, represent the input data for object-1 and object-2. Here, $v_1 \in \mathbb{R}^{T \times C \times H \times W}$, where $T, C, H, W$ represent the time duration, channel, height, and width of the RGB frame, respectively. The audio is denoted as $a_1, a_2 \in \mathbb{R}^{T \times F}$, where $T$ and $F$ denote the time duration and frequency of the audio signal, respectively. The task involves selecting the most fitting object from the video inputs to answer questions (i.e., $q$) according to physical commonsense. The predicted answer is $\hat{Y}$, while the ground-truth answer is $Y$. During the pre-processing phase, the extracted features of audio, video, and question are denoted as $X^a$, $X^v$, and $X^t$, respectively. Here, $X^a, X^t \in \mathbb{R}^d$ refer to the audio and question text features captured as non-sequential data, with $d$ signifying the feature dimension. Conversely, the video feature, represented as sequential data, is denoted as $X^v=\{X^v_1, X^v_2, \cdots, X^v_T\}$, where $T$ indicates the number of video frames and $X^v_i \in \mathbb{R}^d$. Furthermore, we assume that modality data may be missing during both the training and testing phases to address the robustness challenge in physical audiovisual commonsense reasoning. For a given mini-batch of data, $B$ is the batch size. For example, the proportion of missing data in the object-1's video sample is denoted by ${\alpha}_{v_1}$, indicating that ${\alpha}_{v_1} \cdot B$ samples in the mini-batch are missing. This process is similarly applied to the missing of object-2 video or audio data.

\section{Proposed Approach}

\subsection{Overview}
\label{sec:over}

As depicted in Figure~\ref{fig: an overview of model}(a), our proposed method extracts features from input videos and audios using respective encoders, then employs a Disentangled Sequence Encoder~(Sec.~\ref{subsec:dis} and Sec.~\ref{sec: DSE+}) to separate static and dynamic factors. The Counterfactual Learning Module~(Sec.~\ref{subsec:cou}) generates raw and intervened multi-modal features, which are integrated with the question feature. The final predictions are optimized based on the fusion features. To improve the robustness of physical knowledge learning, we developed an enhanced model, RDCL, by introducing an incomplete multi-modal learning module~(IMLM)~(Sec.\ref{Incomplete modality learning Module}) to compensate for missing modalities, as shown in Figure.~\ref{fig3}.

\subsection{Disentangled Sequential Encoder}
\label{subsec:dis}

As shown in Figure~\ref{fig: an overview of model}(b), Disentangled Sequential Encoder~(DSE) is designed to separate static and dynamic factors within multi-modal data. This model enhances the traditional sequential variational auto-encoder~(VAE) by integrating a mutual information term to amplify the disentanglement effect. Specifically, we postulate that the latent representations of the input video's feature, denoted as $X^v_{1:T}$, can be partitioned into a static factor $s$ and dynamic factors $z_{1:T}$, where $z_t$ signifies the latent dynamic representation at the time step $t$. Following~\cite{bai2021contrastively}, we propose that these two factors are mutually independent, expressed as $p(s,z_{1:T})=p(s)p(z_{1:T})$, where $p(\cdot)$ symbolizes the probability distribution. Furthermore, $z_i$ is contingent on $z_{<i}=\{z_0,z_1,..,z_{i-1}\}$, with $z_0=0$, and the reconstruction~\footnote{To simplify, we will use $x_i$ to denote $X^v_{i}$ in the subsequent discussion.} of $x_i$ is independent of other frames given $z_i$ and $s$. Consequently, we aim to learn a posterior distribution $q(z_{1:T},s|x_{1:T})$ where the two factors are disentangled, as expressed in the following equation:
\begin{equation}
\begin{split}
    q(z_{1:T},s|x_{1:T}) &= q(s|x_{1:T})q(z_{1:T}|x_{1:T}) \\
&= q(s|x_{1:T})\prod_{i=1}^T q(z_i|z_{<i}, x_{\leq i}).
\end{split}
\label{eq.2}
\end{equation}

Specifically, we employ the Bi-LSTM~\cite{graves2005framewise} to represent the posterior distribution, where $q(z_i|z_{<i}, x_{\leq i})$ is conditioned on the entire time series by using the hidden states as input, and $q(s|x_{1:T})$ is computed by inputting $x_{1:T}$. Subsequently, we sample the two disentangled factors $s$ and $z_{1:T}$ using the distributions $q(s|x_{1:T})$ and $q(z_i|z_{<i}, x_{\leq i})$ through the reparameterization trick~\cite{kingma2013auto}. 
Afterward, we employ the extracted disentangled factors to reconstruct $x_{1:T}$ using a VAE-based decoder~\cite{bai2021contrastively}. The priors of the static factor $s$ and dynamic factor $z_i$ are defined as Gaussian distributions with $\mathcal{N}(0, I)$ and $\mathcal{N}(\mu(z_{<i}),\sigma^2(z_{<i}))$ respectively, where $\mu(\cdot)$ and $\sigma(\cdot)$ are modeled by Bi-LSTM. The following factorization can formalize the reconstruction process:
\begin{equation}
p(x_{1:T},s,z_{1:T})=p(s)\prod_{i=1}^T p(z_i|z_{<i})p(x_i|z_i,s).
\end{equation}

Furthermore, we incorporate mutual information~(MI) to promote exclusivity between the disentangled factors~({i.e.}, static and dynamic factors) and integrate non-parametric contrastive estimation into the standard loss function for learning latent representations, which can be formulated as:
\begin{equation}
    \mathcal{C}(z_{1:T})= \mathbb{E}_{p_{D}}log \frac{\phi(z_{1:T},x^+_{1:T})}{\phi(z_{1:T},x^+_{1:T})+\sum^{n}_{j=1}\phi(z_{1:T},x^j_{1:T})},
    \label{mutual information_c}
\end{equation}

\noindent where $x^+$ denotes a `positive' sample containing the same object, while $x^j$ $(j=\{1,2,...,n\})$ signifies $n$ `negative' sample with different objects. To counter high dimensionality~\cite{khosla2020supervised}, we employ $\phi(z_{1:T},x^+_{1:T})=exp(sim(z_{1:T},x^{+}_{1:T})/\tau)$, where $sim(\cdot,\cdot)$ signifies the cosine similarity function and $\tau=0.5$ is a temperature parameter. $\mathcal{C}(s)$ can be computed similarly. To construct the `positive' sample, following~\cite{bai2021contrastively}, we adopt content augmentation by randomly rearranging the video's time steps and motion augmentation via Gaussian blur~\cite{chen2020simple}. The results can be denoted as $\mathcal{C}(z^m_{1:T})$ and $\mathcal{C}(s^c)$, where $z^m_{1:T}$ and $s^c$ represent the augmented data of $z_{1:T}$ and $s$, respectively. The Mutual Information~(MI) term $I(\cdot)$ can be expressed as follows:
\begin{align}
    I(z_{1:T};x_{1:T}) &\approx \frac{1}{2}(\mathcal{C}(z_{1:T})+\mathcal{C}(z^m_{1:T})), \\
    I(s;x_{1:T}) &\approx \frac{1}{2}(\mathcal{C}(s)+\mathcal{C}(s^c)).
\end{align}

The objective function can be formulated by adding MI terms to the standard evidence lower bound~(ELBO):
\begin{align}
    \mathcal{L}_{DSE} &= -\text{log}(p(x_{1:T}|z_{1:T})) +\gamma \cdot \left( \mathcal{L}_{KL_s}+ \mathcal{L}_{KL_z}\right) \nonumber \\
    &\quad - \gamma \cdot \left(  I(z_{1:T}; x_{1:T}) + I(s; x_{1:T})\right) + \theta \cdot I(z_{1:T}; s),
    \label{DSE_LOSS}
\end{align}
where
\begin{align}
    \mathcal{L}_{KL_s} &= KL(q(s|x_{1:T}) || p(s)), \\
    \mathcal{L}_{KL_z} &= \sum_{t=1}^T KL(q(z_t|x_{\leq t}) || p(z_t|z_{<t})).
\end{align}
where $\gamma$, $\alpha$, and $\theta$ are hyper-parameters. The complete proof can be found in our supplementary materials.

\subsection{Disentangged Sequential Encoder+}
\label{sec: DSE+}
As detailed in Sec.~\ref{subsec:dis}, Disentangled Sequential Encoder (DSE) was initially proposed in our conference paper~\cite{lv2024disentangled}. To further enhance the model in extracting distinguishing features between the given two objects, we propose a new Disentangled Sequential Encoder+ (DSE+) by improving the selection strategy for `negative' samples in Eq.~\ref{mutual information_c}. Specifically, we incorporate features of the input paired object as additional negative samples. The dynamic factor \(z \in \mathbb{R}^d\) is extracted from the last cell output of the Bi-LSTM (Eq.~\ref{eq.2}), while the static factor \(s \in \mathbb{R}^d\) represents time-invariant features, where \(d\) denotes the feature dimensionality. Departing from prior work~\cite{bai2021contrastively, zhu2020s3vae} by leveraging content-/motion-augmented samples as positive samples, our approach explicitly emphasizes the inherent dissimilarity between dynamic and static factors within object pairs. Then we propose dual contrastive losses to amplify this distinction:
\begin{equation}
    \mathcal{L}_{contra_s} = \max\left(0, \text{sim}(s_1, s_2) - \delta\right), 
    \label{DSE+_1}
\end{equation}
\begin{equation}
    \mathcal{L}_{contra_z} = \max\left(0, \text{sim}(z_1, z_2) - \delta\right),
    \label{DSE+_2}
\end{equation}

\noindent where $s_1, s_2$ and $z_1, z_2$ denote static and dynamic factors of paired objects, respectively. The cosine similarity function $\text{sim}(\cdot, \cdot)$ quantifies feature alignment, while the margin $\delta$ controls the separation threshold between factors. Hence the dual contrastive losses of DSE+ can be incorporated into Eq.~\ref{DSE_LOSS}, formulated as:
\begin{equation}
\mathcal{L}_{DSE+}=\mathcal{L}_{DSE}+\mathcal{L}_{{contra}_s}+\mathcal{L}_{{contra}_z}.
\label{Eq.DSE+ LOSS}
\end{equation}

\subsection{Counterfactual Learning Module}
\label{subsec:cou}

In this section, the static and dynamic factors extracted by DSE/DSE+ are then employed to establish relationships based on physical knowledge in conjunction with the associated audio features. Concurrently, we implement counterfactual relation intervention to enhance the process of knowledge learning.

\subsubsection{Physical Knowledge Relationship}
Inspired by Knowledge Graph~\cite{ding2022mukea}, we posit that the physical knowledge embedded in diverse samples may exhibit certain correlations. Consequently, we propose to model these implicit relationships via a graph structure, and we construct an affinity matrix $A$ to represent these physical knowledge relationships among various objects. Similarly, we create an affinity matrix for audio features and other modalities, resulting in an augmented matrix $ A_X $ defined as follows:
\begin{equation}
A_X = \left[\begin{array}{c:c:c}
   A_{X^a} & A_{X^v_s} & A_{X^v_z}
\end{array}\right],
\end{equation}
where $A_X$ signifies the augmented matrix composed of three affinity matrices. 
With the well-structured affinity matrix $A$, we can augment the video static and dynamic factors, as well as audio features, denoted as $X^v_s$, $X^v_z$, and $X^a$, by facilitating message passing and transfer across different samples, as follows:
\begin{align}
    X &= \left[\begin{array}{c:c:c}
   X^a  & X^v_s & X^v_z\\
\end{array}\right], \\
    F &= A_X \cdot X^\top,
\end{align}
where $F$ represents the transferred features, and $\top$ indicates the transpose of a matrix. By concatenating these three components and passing them through an MLP, we obtain the fused feature $F_1$ and $F_2$ corresponding to object-1 and object-2, respectively. To compute $A$, we use $A_{X^v_s}$ as an example. Firstly, we calculate the similarity matrix $\mathcal{S}$ based on the static factors, where each element $\mathcal{S}^{i,j} \in \mathcal{S}$~($0 < i, j < B$) can be computed as:
\begin{equation}
\mathcal{S}^{i,j} = \exp\left(\frac{\text{sim}(x_i, x_j)}{\tau}\right), \quad x_i, x_j \in X_s,
\end{equation}
where $\text{sim}(\cdot, \cdot)$ denotes the cosine similarity, and $\tau$ is the temperature coefficient. To eliminate the noisy relationships, we apply a near neighbor selection function $\mathcal{T}(\cdot, k)$, which retains the top-$k$ values in each row of $\mathcal{S}$, resulting in a refined matrix $\mathcal{S}^\prime$:
\begin{equation}
\mathcal{S}^\prime = \mathcal{T}(\mathcal{S}, k).
\end{equation}
Finally, we normalize the affinities using the Laplacian matrix $D^{-1}$ of $\mathcal{S}^\prime$, yielding:
\begin{equation}
A_{X^v_s} = D^{-1} \cdot \mathcal{S}^\prime.
\end{equation}
Following a similar calculation, we can obtain $A_{X^v_z}$ and $A_{X^a}$.

\subsubsection{Counterfactual Relation Intervention}

To provide additional supervision for the affinities $A_X$, we propose to emphasize the role of the object's physical knowledge relationship during optimization. Initially, we formulate our method as a Structural Causal Model (SCM)~\cite{pearl2018book}, as depicted in Figure~\ref{fig: an overview of model}(c), and subsequently incorporate causal inference into our method. $\hat{Y}$ denotes the final classification output of the model, which is derived by forwarding the input F into the fusion model and classifier:
\begin{equation}
    \hat{Y}_{X,A_X} = CLS(\phi(F_1,F_2,X^t)),
\end{equation}
where $F_1$ and $F_2$ represent the fused visual-audio features of the input pair $v_1$ and $v_2$, respectively, and $X^t$ signifies the feature of the question text. `$CLS$' and `$\phi$' denote the classifier and fusion model, respectively, with further details provided in the following Section~\ref{Sec: Optimization}. The process of generating the output $\hat{Y}$ from the input $X$ can be considered as two types of effects: a direct effect $X \rightarrow \hat{Y}$, and an indirect effect $X \rightarrow A_X \rightarrow \hat{Y}$. Our final loss function aims to maximize the likelihood estimation of $\hat{Y}$, which influences both types of effects in an end-to-end manner, resulting in an insufficient enhancement of $A_X$ in the indirect effects path. Therefore, we employ the Total Indirect Effect (TIE) to emphasize the effect of $A_X$:
\begin{equation}\label{optimizatie for TIE}
    \hat{Y}_{TIE}=\hat{Y}_{X,A_X}-\mathbb{E}_{X^*}[\hat{Y}_{X,A_{X^*}}],
\end{equation}
where $\hat{Y}_{X,A_{X^*}}$ refers to the results calculated by substituting the original affinity $A_X$ with an intervened one $A_{X^*}$, and $X^*$ represents the given intervened inputs. Note that $\hat{Y}_{X,A_{X^*}}$ cannot occur in reality because affinities $A_{X^*}$ originate from $X^*$, which is referred to as counterfactual intervention. Therefore, modifying $\hat{Y}_{X, A_X}$ to $\hat{Y}_{X, A_{X^*}}$ is equivalent to keeping all features constant but only altering the affinity $A_X$. We compute the expectation of that effect to obtain a more stable one, and the intervened input features $X^*$ are sampled by a Gaussian distribution:
\begin{equation}
    X^*=X_\sigma \cdot W + X_\mu,
\end{equation}
where $W$ is a standard random vector with the same dimension as $X$, and both mean $X_\mu$ and standard deviation $X_\sigma$ are learned via the re-parameterization trick.

\begin{figure*}
    \centering
    \includegraphics[width=0.82\linewidth]{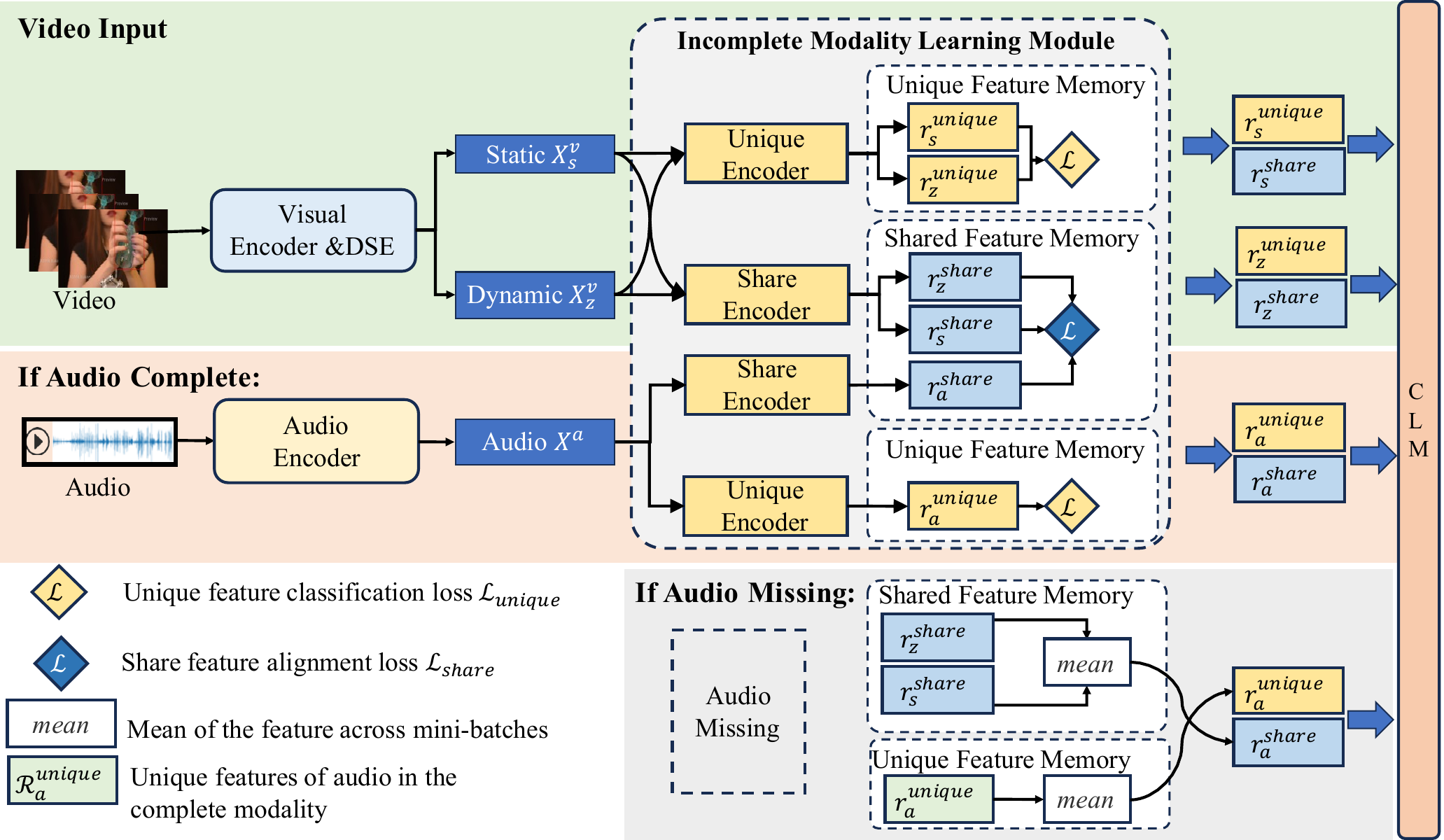}
    \vspace{-3mm}
    \caption{Illustration of our proposed RDCL model. The upper part shows our proposed Incomplete Multi-Modal Learning Method (IMLM) within RDCL during the training stage when the training data is modality-complete. IMLM comprises a unique encoder and a shared encoder, along with a Shared Feature Memory and a Unique Feature Memory. As a plug-in model, the features processed by the IMLM are subsequently fed into the Counterfactual Learning Module (CLM). The lower part presents RDCL during the inference stage when audio data is missing, and we utilize the average value across the shared feature memory to substitute for the missing audio feature.}
    \label{fig3}
    \vspace{-2mm}
\end{figure*}

\subsubsection{Fusion Model and Optimization}
\label{Sec: Optimization}

Our proposed approach functions as a plug-and-play module, capable of seamless integration into various multimodal fusion models. We will illustrate the application of our method using LateFusion method~\cite{pandeya2021deep} as the example, which is based on linear classifiers. For the given object-1, object-2, and the textual feature ($F_1$, $F_2$, and $X^t$), we employ two multilayer perceptrons (MLPs) as the fusion model, expressed as:
\begin{equation}
    \phi(F_1, F_2, X^t)={MLP}_1({MLP}_2(F_1 \|F_2)\|X^t),
\end{equation}
where $\|$ denotes row-wise concatenation, and $\text{MLP}_1$ and $\text{MLP}_2$ represent two independent MLPs with distinct parameters. Subsequently, we employ a fully connected layer as the classifier, with the input dimension of $d$ (the hidden feature dimension of the model) and the output dimension of two, indicating the selection of the suitable object for the input text between the two. For $\mathcal{L}_{TIE}$, we minimize the cross-entropy between $\hat{Y}_{TIE}$ and the corresponding labels $Y_{GT}$, which can be formulated as:
\begin{equation}
    \mathcal{L}_{TIE} = - Y_{GT} \log(\hat{Y}_{TIE}).
\label{Loss TIE}
\end{equation}

Finally, our optimization goal can be formulated as:
\begin{equation}
    \mathcal{L}_{DCL}=\mathcal{L}_{DSE+}+\mathcal{L}_{TIE}.
    \label{Loss_DCL}
\end{equation}

\subsection{Incomplete Multi-Modal Learning Module}
\label{Incomplete modality learning Module}
In this section, we introduce an Incomplete Multi-Modal Learning Module~(IMLM) to address the challenge of missing modalities in real-world applications. The proposed IMLM aims to investigate the unique and shared semantic information among video static factors, dynamic factors, and audio features. The shared semantic information is subsequently leveraged to compensate for any missing modalities. As illustrated in Fig.~\ref{fig3}, the architecture of IMLM is divided into two components: complete modalities~(Sec.~\ref{Model Structure of Complete Modalities}) and missing modalities~(Sec.~\ref{Model Structure of Missing Modalities}).

\subsubsection{Complete Modalities Learning}
\label{Model Structure of Complete Modalities}

According to the proposed DSE/DSE+, for a given object $i$, we denote its input audio features as $x^a_i \in X^a$, while $x^z_i \in X^v_z$ and $x^s_i \in X^v_s$ represent the dynamic and static factors, respectively. Since all subsequent descriptions pertain to object $i$, the subscript $i$ is omitted hereafter.

For samples with complete modalities, our objective is to extract both the shared semantic information, which encapsulates the physical properties of the object, and the unique semantic information, which captures the distinct characteristics of each modality. These features are encoded using a shared feature encoder $f_{share}(\cdot)$ and a unique feature encoder $f_{unique}(\cdot)$, formulated as follows:
\begin{align}
    r^{share}_m &= f_{share}(x^m), \quad m \in \{a, z, s\}, \\ \label{eq:shared_unique_representation}
    r^{unique}_m &= f_{unique}(x^m), \quad m \in \{a, z, s\},
\end{align}
where $r^{share}_m$ and $r^{unique}_m$ denote the shared and unique features of modality $m$, respectively. Both $f_{share}(\cdot)$ and $f_{unique}(\cdot)$ are implemented as independent two-layer MLPs, with parameters shared across all modalities.
Subsequently, the shared features $r^{share}_m$ of complete modalities are stored in memory.

Then we project the concatenated shared and unique features of each modality into a latent feature space with the same dimensionality as the original features $x^a$ and $x^v$. This projection is achieved using an MLP denoted as $f_{\text{pro}}(\cdot)$. Inspiring from residual connections~\cite{he2016deep}, we incorporate the projected features into the original features through addition, as expressed by:
\begin{equation}
    x^{m'} = f_{pro}(r^{share}_m \| r^{unique}_m) + x^m, 
    \label{eq:share_unique_projection_1}
\end{equation}
where $\|$ denotes concatenation, and $x^{m}$ and $x^{m'}$ represent the features before and after processing by the IMLM, respectively, and $m \in \{a, z, s\}$. The resulting features $x^{m'}$ are then utilized as the static, dynamic, and audio features for the subsequent Counterfactual Learning Module.

\subsubsection{Missing Modalities Learning}
\label{Model Structure of Missing Modalities}

In scenarios involving missing modalities, the proposed IMLM is designed to mitigate semantic information loss in physical attributes. Specifically, we reconstruct the missing modality's semantic features by leveraging shared information across static, dynamic, and audio modalities. For instance, in cases where audio data is missing\footnote{In this paper, we use missing audio scenarios as an illustrative case.} at a rate of $\alpha_a$, we define the set of missing data $B_{miss}$ for a batch of size $N$ as follows:
\begin{equation}
    B_{{miss}} = \{b_1, b_2, \dots, b_{N \times \alpha_a}\},
\end{equation}
where $b_i$ denotes the $i$-th missing data point, and $1 \leq i \leq N \times \alpha_a$. Subsequently, we use $B_{{com}}$ to represent the complete set of modalities, where $N = \|B_{{miss}}\| + \|B_{{com}}\|$. For the audio data $a_i \in a_1$ of object-1, it can be represented as:
\begin{equation}
    a_i = \begin{cases} 
0, & \text{if } i \in B_{{miss}} \\
a_i, & \text{if } i \in B_{{com}}.
\end{cases}
\end{equation}

The same operation can be applied to object-1's and object-2's audio data, as well as the corresponding video data.

Subsequently, we employ the DSE/DSE+ to encode and decouple the video, while using an audio encoder to encode the audio. For the available modalities, we extract shared and unique features by Eq.~\ref{eq:shared_unique_representation}, denoted as $r^{share}_m$ and $r^{unique}_m$ respectively, where $m \in \{a, z, s\}$. For samples in the missing subset $B_{miss}$, represented as $\{{x_i^a},  i \in B_{miss}\}$, we directly utilize the shared features of the corresponding modalities (e.g., static factor $r^{share}_{s, i}$ and dynamic factor $r^{share}_{z, i}$) as the audio shared features:
\begin{equation}
    r^{share}_{a, i} = \frac{1}{2} \left( r^{share}_{z, i} + r^{share}_{s, i} \right), \quad i \in B_{miss}.
\end{equation}

For unique features, we compute them as the mean of the unique features from other samples:
\begin{equation}
    {r}^{unique}_{a, i} = \frac{1}{\|B_{com}\|} \sum_{j=1}^{\|B_{com}\|} {r}^{unique}_{a, j}, \quad i \in B_{miss}, \, j \in B_{com}.
\end{equation}

Finally, we project the shared and unique features derived from static factor, dynamic factor, and audio into their respective feature spaces using Eq.~\ref{eq:share_unique_projection_1}.

\subsubsection{Optimization}
Our method maintains an identical architecture during both training and testing. During training, we employ a domain classification objective to optimize the features extracted by the unique encoder. These features are classified using the classifier \(f_{modal}(\cdot)\), and the unique feature classification loss is computed via cross-entropy\footnote{Here, we use the linear classifiers employed in late fusion as an example.}:
\begin{equation}
    \hat{Y}_{unique} = f_{modal}(r^{unique}_m), \quad m \in \{a, s, z\},
\end{equation}
\begin{equation}
\begin{split}
    \mathcal{L}_{unique} = -\frac{1}{\|B_{com}\|} \sum_{i=1}^{\|B_{com}\|} \left[ Y_{unique,i} \log(\hat{Y}_{unique,i}) \right. \\
    \left. + (1 - Y_{unique,i}) \log(1 - \hat{Y}_{unique,i}) \right],
\end{split}
\end{equation}
where \(Y_{unique}\) and \(\hat{Y}_{unique}\) denote the ground truth and predicted values, respectively. \(Y_{unique} = 1\) indicates \(r^{unique}_m\) belongs to modality \(m\); otherwise, \(Y_{unique} = 0\). 

To align the shared semantic information encapsulating the physical properties of objects,  we enforce consistency in the shared feature representations $r^{share}_m$ ($m \in \{a, z, s\}$) by minimizing the symmetrized L1 loss between all modality pairs. The share feature alignment loss is defined as:
\begin{equation}
\begin{aligned}
    \mathcal{L}_{share} &= \sum_{m \neq m'} \| r^{share}_m - r^{share}_{m'} \|_1,
\end{aligned}
\end{equation}
where \( r^{share}_m \) and \( r^{share}_{m'} \) represent shared features of \( m \) and \( m' \), respectively, and \( \| \cdot \|_1 \) denotes the L1 norm.  

The overall loss function of the model under incomplete modalities is defined as:
\begin{equation}
\mathcal{L}_{IMLM} = \mathcal{L}_{unique} + \mathcal{L}_{share}.
\label{IMLM_LOSS}
\end{equation}

Finally, the loss function for our RDCL is formulated as:
\begin{equation}
\mathcal{L}_{RDCL} = \mathcal{L}_{DSE+} + \mathcal{L}_{TIE} + \mathcal{L}_{IMLM}.
\end{equation}

\section{Experiments}

\subsection{Experimental Setup}
\noindent{\bf Dataset.} The Physical Audiovisual CommonSense Reasoning Dataset (PACS)~\cite{yu2022pacs} is a compilation of 13k question-answer pairs curated to assess physical commonsense reasoning abilities. PACS encompasses 1,377 distinct physical commonsense questions covering a range of physical properties, supplemented by 1,526 video and audio clips sourced from YouTube. The PACS dataset comprises 13,400 data points in total, with the PACS-Material subset containing 4,349. In line with~\cite{yu2022pacs}, we segregate PACS into training, validation, and testing sets with 11,044, 1,192, and 1,164 data points respectively, each containing 1,224, 150, and 152 objects respectively. The PACS-Material subset is partitioned into 3,460, 444, and 445 data points for training, validation, and testing respectively, maintaining the same object distribution as PACS. To ensure unbiased model evaluation, we assess our method on both the complete dataset and a subset concentrating on material-related issues, presenting the results for each subset separately during testing.

\noindent{\bf Evaluation Metric.}
Following~\cite{yu2022pacs}, we employ accuracy as the evaluation metric for both PACS and PACS-material subsets. All experimental results are reported as the average of five independent runs.

\noindent{\bf Implementation.} Our proposed model is developed using PyTorch and executed on a single NVIDIA RTX 3090 GPU. Specifically, we preprocess each video by downsampling to $T=8$ frames and establish the feature dimension as $d=256$. In the Disentangled Sequence Encoder, a hidden layer size of 256 is utilized for the Bi-LSTM. During the optimization process, we establish a batch size of 64, comprising 64 video pairs and their corresponding questions. The hyperparameters \(\gamma\), and \(\theta\) are assigned values of 1 and 50, respectively. In the Counterfactual Learning Module, $\tau=2$ and $k=5$ are employed for calculating similarities and establishing the physical knowledge relationships. The parameter count for AudioCLIP is 182M, while AudioCLIP with DCL has 192M, and RDCL has 214M. The inference time for AudioCLIP with DCL is 277 seconds. For more details please refer to the supplementary material. 

\noindent{\bf Compared Methods.}~To validate the effectiveness of our proposed approach, we compare it with the following baseline methods: {\bf 1) Late fusion}~\cite{pandeya2021deep} utilizes separate encoders for text, image, audio, and video to extract unimodal features. These features are concatenated and passed through a linear layer to generate multimodal embeddings for prediction. {\bf 2) CLIP/AudioCLIP}~\cite{radford2021learning,guzhov2022audioclip} embeds video, text, and audio data into a shared vector space using CLIP and AudioCLIP. A linear layer is then applied to produce multimodal embeddings for prediction. Note that since CLIP cannot extract audio features, audio data is excluded in experiments involving CLIP. {\bf 3) UNITER}~\cite{chen2020uniter} is a pre-trained model for image and text that has been trained on four image-text tasks and has demonstrated strong performance on tasks such as NLVR2~\cite{suhr2019corpus}. {\bf 4) MLLMs.} To evaluate the performance of existing large models on physical commonsense reasoning, we test popular models, including Gemini~\cite{team2024gemini} and GPT-4V~\cite{achiam2023gpt}, as well as the open-source model Qwen-VL~\cite{bai2023qwen}. For all the aforementioned benchmark methods, we adhere to the parameters reported in their respective papers.

\begin{table}[t]
\caption{Quantitative results comparing baseline methods with our proposed method.}
\vspace{-2mm}
\begin{adjustbox}{width=0.49\textwidth}
\begin{tabular}{lcccc}
\toprule
\multirow{2}{*}{Baseline Model} & \multicolumn{3}{c}{Accuracy (\%)} &     \\
                                & PACS        & $\Delta$   & PACS-Material & $\Delta$   \\ \hline \rowcolor{gray!10}
Gemini~\cite{team2024gemini}                          & 65.7  & -   & -    & -   \\ \rowcolor{gray!10}
Qwen-VL~\cite{bai2023qwen}                            & 55.7  & -   & -    & -   \\ \rowcolor{gray!10}
GPT-4V~\cite{achiam2023gpt}                          & 51.3  & -   & -    & -   \\ \hline
Late Fusion~\cite{pandeya2021deep}                     & 55.0 ± 1.1  & -   & 67.4 ± 1.5    & -   \\
Late Fusion~\cite{pandeya2021deep} w/ DCL              & \underline{57.7 ± 0.9}  & \underline{+2.7} & \underline{69.7 ± 1.2}    & \underline{+2.3}
 \\ 
Late Fusion~\cite{pandeya2021deep} w/ DCL (DSE+)           & {\bf 58.1 ± 0.8}        & {\bf +3.1} & {\bf 70.6 ± 1.1}        & {\bf +3.2} \\ \hline
CLIP~\cite{radford2021learning}                            & 56.3 ± 0.7  & -   & 72.4 ± 1.1    & -   \\
CLIP~\cite{radford2021learning}  w/ DCL                     & \underline{58.4 ± 0.8}  & \underline{+2.1} & \underline{75.4 ± 1.2}    & \underline{+3.0}   \\ 
CLIP~\cite{radford2021learning}  w/ DCL (DSE+)      & {\bf 60.6 ± 0.7}      & {\bf +2.5} & {\bf 77.5 ± 1.1}         & {\bf +5.1} \\ \hline
UNITER(Large)~\cite{chen2020uniter}                   & 60.6 ± 2.2  & -   & 75.0 ± 2.8    & -   \\
UNITER~\cite{chen2020uniter} w/ DCL                   & \underline{62.0 ± 2.4}  & \underline{+1.4} & \underline{75.7 ± 2.8}    & \underline{+0.7} \\ 
UNITER~\cite{chen2020uniter} w/ DCL (DSE+)                       & {\bf 62.7 ± 2.1}       & {\bf +2.1} & {\bf 76.6 ± 2.5}         & {\bf 1.6} \\ \hline
AudioCLIP~\cite{guzhov2022audioclip}                       & 60.0 ± 0.9  & -   & 75.9 ± 1.1    & -   \\
AudioCLIP~\cite{guzhov2022audioclip}  w/ DCL                & \underline{63.2 ± 0.8}  & \underline{+3.2} & \underline{76.2 ± 1.4}    & \underline{+0.3} \\ 
AudioCLIP~\cite{guzhov2022audioclip}  w/ DCL (DSE+)                  & {\bf 65.3 ± 1.2}       & {\bf +5.3} & {\bf 79.7 ± 1.5}         & {\bf +3.8} \\ \bottomrule
\end{tabular}
\end{adjustbox}
\label{main table}
\vspace{-3mm}
\end{table}

\begin{figure*}[ht]
    \centering
    \includegraphics[width=0.88\linewidth]{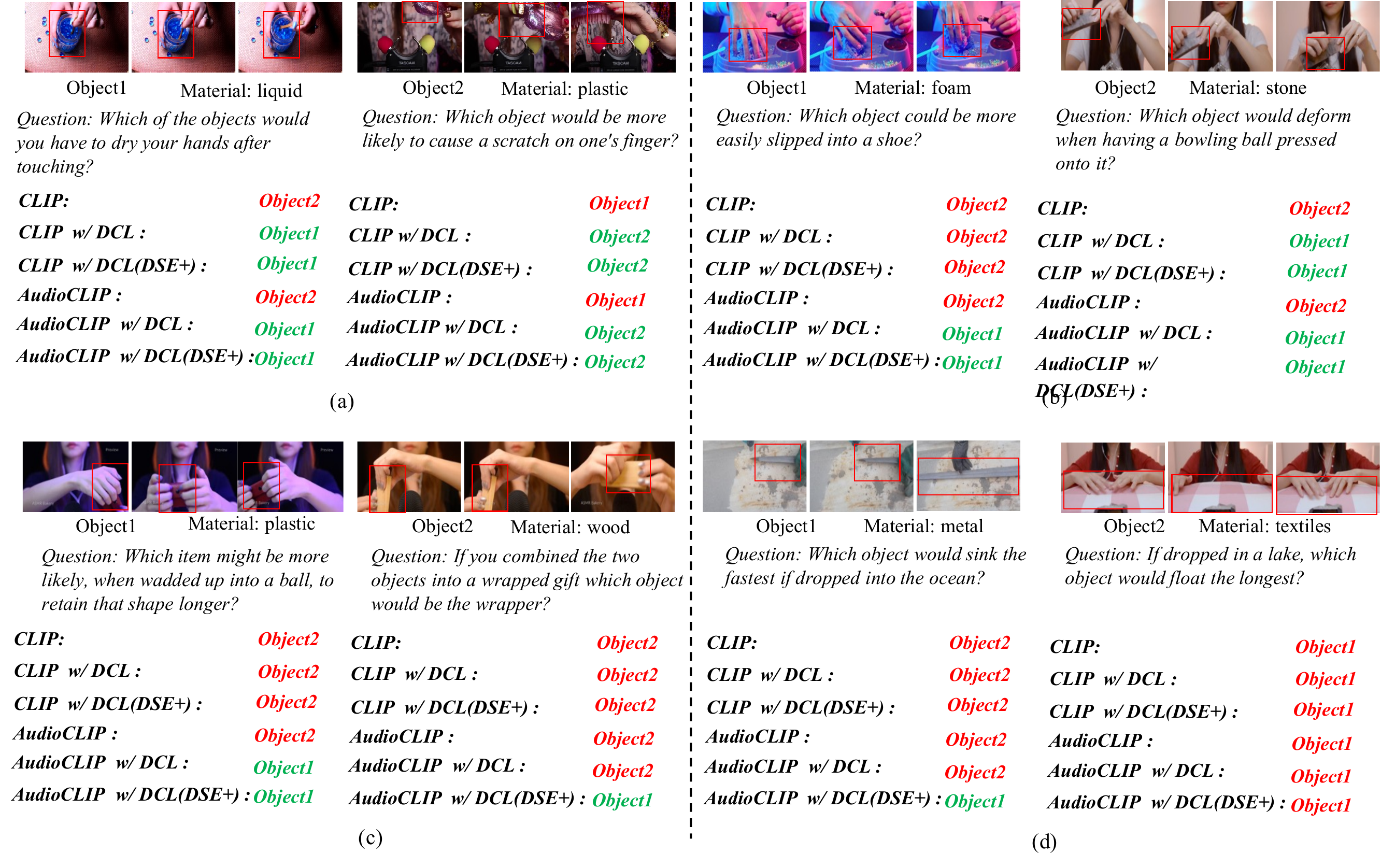}
    \vspace{-3mm}
    \caption{Qualitative Results of baseline w/ and w/o our proposed method, where `Material' refers to the material of the object. The correct answers are depicted in green while the incorrect ones are depicted in red.}
    \label{qualitative}
    \vspace{-3mm}
\end{figure*}

\subsection{Comparison to Baselines}

\textbf{Quantitative Results.}
We present quantitative performance comparisons on the PACS dataset in Table~\ref{main table}. The results demonstrate that integrating our proposed DCL method leads to consistent improvements across all baseline models. Specifically, Late Fusion and UNITER achieve absolute accuracy gains of 2.7\% and 1.4\%, respectively. Similarly, CLIP and AudioCLIP, which align image, audio, and text modalities into a shared embedding space, show improvements of 2.1\% and 3.2\%, respectively. These results underscore the strong reasoning and generalization capabilities of our DCL approach.
These results highlight the strong reasoning and generalization capabilities of our DCL method. 
Furthermore, we evaluate the enhanced variant, DCL with DSE+, which introduces contrastive losses for static and dynamic factors. As shown in Table~\ref{main table}, DCL (DSE+) yields additional performance gains over DCL across all baselines. For instance, UNITER and AudioCLIP achieve absolute improvements of 0.7\% and 2.1\%, respectively, highlighting the effectiveness of the proposed contrastive losses in refining feature representations.
Notably, even with DCL, CLIP 's performance remains below that of AudioCLIP, emphasizing the importance of audio information in physical commonsense reasoning. When comparing CLIP and AudioCLIP enhanced with DCL, the inclusion of audio information results in a significant absolute improvement of 4.8\%. However, with the further enhancements of DCL (DSE+), CLIP achieves an accuracy of 60.6\%, matching the performance of AudioCLIP. This suggests that CLIP, despite lacking audio information, can achieve comparable results to AudioCLIP when equipped with DSE+, demonstrating the DSE+'s ability to effectively handle video features. The same trend can be observed on the PACS-Material dataset, where our method consistently enhances material reasoning performance across all models. This indicates that our approach serves as a versatile, plug-and-play module that can be seamlessly integrated into various architectures to improve their reasoning capabilities. Especially, experiments with current multimodal large language models (MLLMs) on the PACS benchmark reveal that Qwen-VL and GPT-4V exhibit significant performance gaps compared to both the baseline and the baseline w/ DCL. Notably, our AudioCLIP w/ DCL (DSE+) achieves performance approaching that of Gemini, demonstrating the DCL's superiority on this benchmark. Finally, it is worth noting that all objects in the test set were excluded from the training and validation sets, showcasing the zero-shot reasoning ability of our model. This further validates the generalizability of our proposed method.

\textbf{Qualitative Results.}~ Figure~\ref{qualitative} presents comparative visualization results for identical questions. In Figure~\ref{qualitative}(a), both objects are small in size, but Object-1 exhibits a deformable, time-varying shape. Our DCL model accurately captures Object-1’s liquid-like, dynamically mutable properties, enabling consistent correct predictions across both questions by leveraging this distinctive characteristic. Figure~\ref{qualitative}(b) highlights DCL’s capacity to model physical knowledge embedded in audio data. Since Object-1 emits a foam-like acoustic signature distinct from Object-2, CLIP which relies solely on visual data—fails to resolve Question-1 correctly. By contrast, AudioCLIP, augmented with auditory input, achieves the correct prediction. However, audio-only approaches remain error-prone: in question-2, accurate reasoning requires synthesizing both auditory features (e.g., sound texture) and dynamic visual cues (e.g., small size). While AudioCLIP falters due to insufficient motion modeling, our DCL integrates multimodal dynamics to maintain robustness. Figure~\ref{qualitative}(c) demonstrates a critical edge case where the two objects share nearly identical geometries and manual interaction patterns. Baseline models fail here, but our DCL (DSE+) resolves question-1 by exploiting audio-derived material plasticity cues (e.g., distinguishing plastic deformability from wood rigidity). The second question requires modeling the comparative relationship between the two objects, where DCL (DSE+) uniquely succeeds through contrastive binary loss optimization. This approach explicitly guides attention to pairwise physical property interactions, proving effective for object pair comparative tasks. Figure~\ref{qualitative}(d) illustrates a failure case, where the input video of Object-1 simultaneously contains two distinct objects (a knife and a stone). All models struggle to determine which object’s characteristics are being queried, leading to incorrect predictions. This exceptional case stems from dataset limitation rather than the models' design, and adopting the advanced object detection model can mitigate the issue. More results refer to the supplementary material.

\subsection{Ablation Study}
In this section, we conduct ablation studies to evaluate the contribution of each module in our proposed method.

\begin{figure*}
    \centering
    \includegraphics[width=1.0\linewidth]{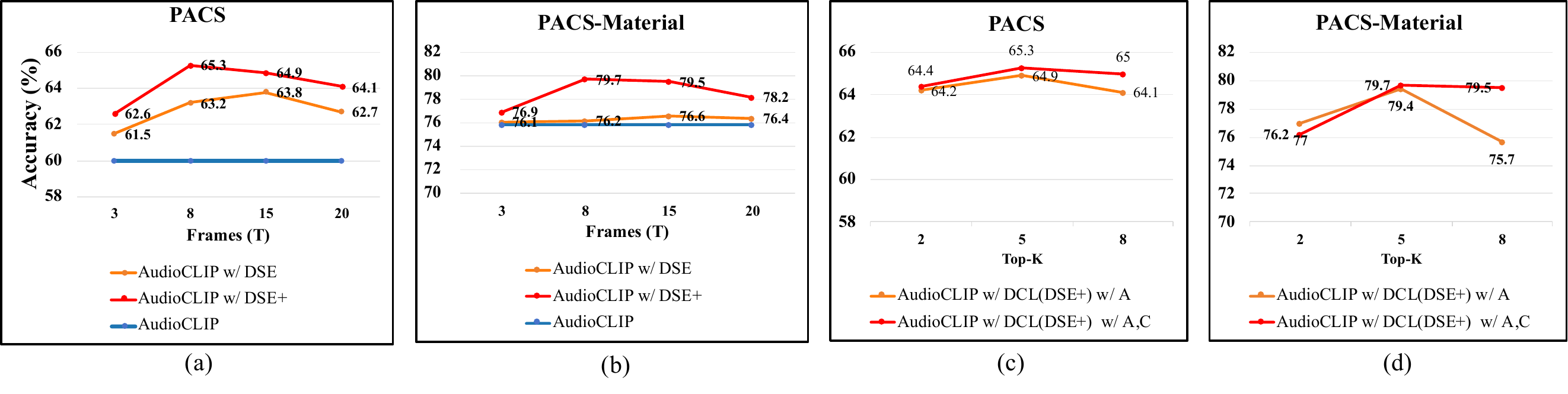}
    \vspace{-8mm}
    \caption{Performance comparison of various hyperparameters. Figures (a) and (b) show the performance of AudioCLIP with different frame lengths \( T \) in DSE and DSE+ on the PACS and PACS-Material datasets. Figures (c) and (d) illustrate the performance of AudioCLIP with varying numbers of top-\( K \) physical knowledge relationships on the same datasets.}
    \label{fig: T of DSE}
    \vspace{-3mm}
\end{figure*}

\subsubsection{DSE and DSE+}
The Disentangled Sequential Encoder (DSE) is designed to decompose the sampled video features into two distinct components: static factors and dynamic factors. To evaluate the effectiveness of DSE and DSE+, we analyze their performance from three perspectives: The independent application of DSE and DSE+, the number of frames to disentangle, and the impact of disentanglement on the CLM.

\noindent{\bf The independent application of DSE and DSE+.}~Table~\ref{DSE} presents a performance comparison between our proposed Disentangled Sequential Encoder (DSE), its enhanced version (DSE+), and various baseline models. As illustrated in rows 3, 7, and 11 of Table~\ref{DSE}, the integration of DSE leads to absolute performance improvements of 1.2\%, 1.0\%, and 1.1\% for the three baselines, respectively, with AudioCLIP achieving the highest performance. A similar trend is observed in the PACS-Material subset. Compared to the approach in rows 2, 6, and 10, which employs a Multi-Layer Perceptron (MLP) with the same number of parameters as DSE, the incorporation of DSE results in enhanced accuracy across both problem subsets. This demonstrates that DSE effectively improves the representation of physical characteristics in video features.
Furthermore, the adoption of DSE+ further enhances baseline performance. As shown in rows 4, 8, and 12 of Table~\ref{DSE}, DSE+ achieves absolute improvements of 2.5\%, 2.5\%, and 3.7\%, respectively, across the three baselines, compared to the non-decoupling approach. Additionally, when compared to DSE, DSE+ yields absolute accuracy gains of 1.3\%, 1.5\%, and 2.5\% on the PACS dataset. These improvements are also consistent in the PACS-Material subset. In contrast to previous methods that relied solely on contrastive learning within sample features, incorporating binary contrastive losses enhances the distinctiveness between object pairs, further improving the model's accuracy.

{\bf The number of frames to disentangled.}~We extract continuous dynamic features and consistent static features from the sampled $T$ video frames. Intuitively, a larger number of sampled frames is expected to enhance the disentanglement performance. To evaluate this, we conducted experiments under four conditions with varying numbers of frames ($T=3, 8, 15, 20$), as summarized in Figure~\ref{fig: T of DSE}.  The results demonstrate that the accuracy of both PACS and PACS-material peaks at $T=8$. Notably, increasing the number of frames to $T=15$ or $T=20$ does not yield further improvements in accuracy, despite the associated increase in computational cost.  Conversely, the lowest performance is achieved when $T=3$, indicating that an insufficient number of frames adversely affects the disentanglement process. Hence the number of sampled frames impacts the final disentanglement performance, with both excessively high and low values leading to suboptimal results.

\begin{table}[tbp]
\caption{Quantitative results of baselines with our DSE and DSE+. MLP denotes a fully connected layer with the same number of parameters as DSE.}
\vspace{-2mm}
\label{DSE}
\begin{adjustbox}{width=0.49\textwidth}
\begin{tabular}{lcccc}

\hline
\multicolumn{1}{c}{\multirow{2}{*}{Baseline Model}} & \multicolumn{3}{c}{Accuracy (\%)}  &     \\
\multicolumn{1}{c}{}                                & PACS        & $\Delta$    & PACS-Material & $\Delta$   \\ \toprule
Late Fusion\cite{pandeya2021deep}                                           & 55.0 ± 1.1  & -    & 67.4 ± 1.5    & -   \\
Late Fusion\cite{pandeya2021deep}   w/ MLP                                  & 54.9 ± 0.9  & -0.1 & 67.7 ± 1.1    & +0.3   \\
Late Fusion\cite{pandeya2021deep}   w/ DSE                               & \underline{56.2 ± 0.8}  & \underline{+1.2}  & \underline{68.5 ± 1.2}    & \underline{+0.9} \\ 
Late Fusion\cite{pandeya2021deep}   w/ DSE+                             & {\bf 57.5 ± 0.7}       & {\bf +2.5}  & {\bf 69.1 ± 1.1}        & {\bf +1.7} \\ \hline
CLIP~\cite{radford2021learning}                                                & 56.3 ± 0.7  & -    & 72.4 ± 1.1    & -   \\
CLIP~\cite{radford2021learning} w/ MLP                                         & 56.5 ± 0.5  & +0.3  & 72.6 ± 1.2    & +0.2  \\
CLIP~\cite{radford2021learning} w/ DSE                                      & \underline{57.0 ± 0.6}   & \underline{1.0}    & \underline{73.2 ± 1.1}    & \underline{0.8} \\ 
CLIP~\cite{radford2021learning} w/ DSE+                                   & {\bf 58.5 ± 0.6}       & {\bf +2.5}  & {\bf 74.1 ± 1.4}         & {\bf +1.7} \\ \hline
AudioCLIP~\cite{guzhov2022audioclip}                                            & 60.0 ± 0.9  & -    & 75.9 ± 1.1    & -   \\
AudioCLIP~\cite{guzhov2022audioclip}  w/ MLP                                    & 60.3 ± 0.8  & +0.3     & 76.2 ± 1.3    & +0.3 \\
AudioCLIP~\cite{guzhov2022audioclip}  w/ DSE                                 & \underline{61.1 ± 0.8}  & \underline{+1.1}  & \underline{76.0 ± 1.0}    & \underline{+1.0}   \\ 
AudioCLIP~\cite{guzhov2022audioclip}  w/ DSE+                               & {\bf 63.7 ± 0.9}       & {\bf +3.7}  & {\bf 78.2 ± 1.4}         & {\bf +3.2} \\ \bottomrule
\end{tabular}
\end{adjustbox}
\vspace{-3mm}
\end{table}

\begin{table}[tbp]
\caption{Ablation study of CLIP and AudioCLIP with DSE, DSE+, Physical Knowledge Relationship~($A$) and Counterfactual Relation Intervention~($C$)}
\vspace{-2mm}
\begin{adjustbox}{width=0.49\textwidth}
\begin{tabular}{lcccc}
\toprule
\multirow{2}{*}{Baseline Model} & \multicolumn{4}{c}{Accuracy (\%)}         \\
& PACS       & $\Delta$    & PACS-Material & $\Delta$     \\ \midrule
CLIP~\cite{radford2021learning}                                                & 56.3  & -    & 72.4     & -     \\
CLIP~\cite{radford2021learning} w/ A                                              & 54.5       & -1.8 & 51.5          & -20.9 \\
CLIP~\cite{radford2021learning} DSE w/ A                                      & \underline{57.8}  & \underline{+1.5}  & \underline{74.5}     & \underline{+2.1}  \\ 
CLIP~\cite{radford2021learning} DSE+ w/ A                                    & {\bf 59.8}       & {\bf +3.5}  & {\bf 76.6}          & {\bf +4.2}   \\ \midrule
CLIP~\cite{radford2021learning} w/ A,C                                            & 56.4       & +0.1  & 68.9          & -3.5  \\
CLIP~\cite{radford2021learning} DSE w/ A,C                                     & \underline{58.4}  & \underline{+2.1}  & \underline{75.4}    & \underline{+3.0}    \\ 
CLIP~\cite{radford2021learning} DSE+ w/ A,C                                   & {\bf 60.6}       & {\bf +3.7}  & {\bf 77.5}          & {\bf +5.1}   \\ \midrule
AudioCLIP~\cite{guzhov2022audioclip}                                           & 60.0  & -    & 75.9    & -     \\
AudioCLIP~\cite{guzhov2022audioclip} w/ A                                         & 59.9       & -0.1 & 70.3          & -5.6  \\
AudioCLIP~\cite{guzhov2022audioclip} DSE w/ A                                 & {\bf 61.9}  & {\bf +1.9}  & {\bf 75.8}     & {\bf -0.1}  \\ 
AudioCLIP~\cite{guzhov2022audioclip} DSE+ w/ A                               & \underline{61.2}       & \underline{+1.2}  & \underline{75.2}          & \underline{-0.7}  \\ \midrule
AudioCLIP~\cite{guzhov2022audioclip} w/ A,C                                       & 60.9       & +0.9  & 75.1            & -0.8      \\ 
AudioCLIP~\cite{guzhov2022audioclip} DSE w/ A,C                                & \underline{63.2}  & \underline{+3.2}  & \underline{76.2}    & \underline{+0.3}  \\ 
AudioCLIP~\cite{guzhov2022audioclip} DSE+ w/ A,C                              & {\bf 65.3}       & {\bf +5.3}  & {\bf 79.7}          & {\bf +3.8}   \\
\bottomrule
\end{tabular}
\end{adjustbox}
\label{table:CLM}
\vspace{-3mm}
\end{table}

{\bf The Impact of Disentanglement on the CLM.}~We investigated whether physical knowledge relationships~(denoted as $A$) could remain effective in the absence of disentangled static and dynamic factors. As shown in Table~\ref{table:CLM}, we conducted experiments by replacing the DSE with an MLP to eliminate the effects of disentanglement while keeping the number of parameters consistent. Without the DSE, establishing physical knowledge relationships between objects using only visual features from CLIP (CLIP w/ A) resulted in accuracy reductions of 1.8\% and 20.9\% for PACS and PACS-Material, respectively, compared to CLIP DSE w/ A. In contrast, incorporating the DSE (CLIP DSE w/ A) led to accuracy improvements of 3.1\% and 23.0\% for PACS and PACS-Material, respectively. This not only mitigated the performance degradation associated with introducing physical knowledge relationships but also enhanced the baseline performance. A similar trend was observed in AudioCLIP, where the AudioCLIP DSE w/ A improved accuracy by 2.3\% and 4.4\% in the two subsets, respectively, compared to AudioCLIP w/ A. However, in the case of AudioCLIP on the PACS-Material subset, the AudioCLIP DSE w/ A still underperformed compared to the standard AudioCLIP, highlighting the importance of counterfactual interventions in material-related tasks.

\begin{figure*}[]
    \centering
    \includegraphics[width=0.9\linewidth]{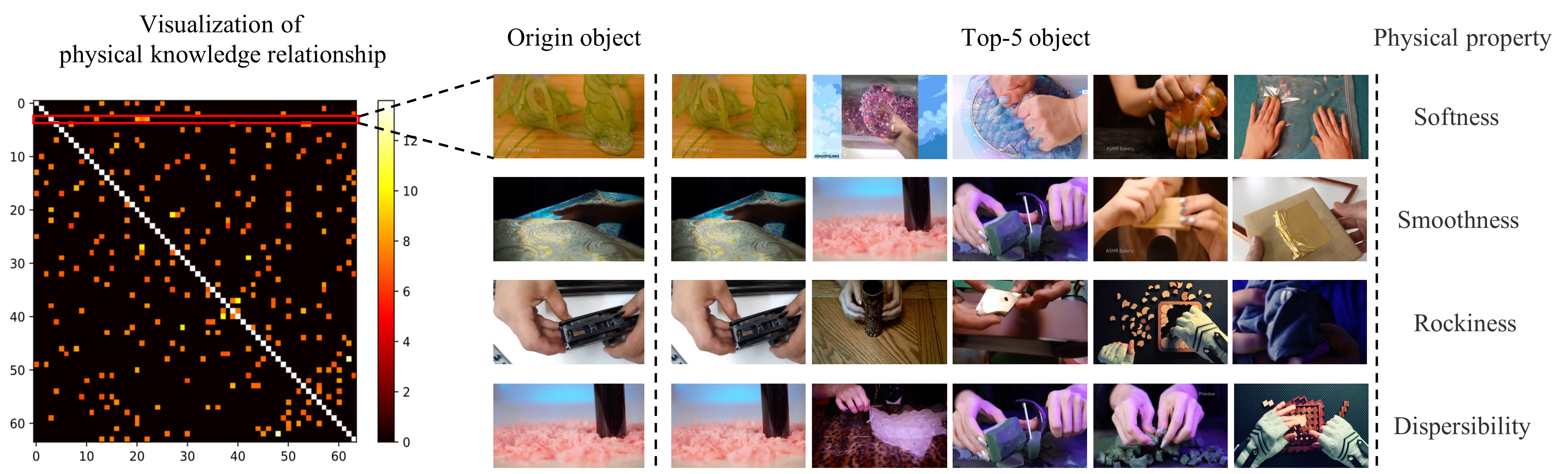}
    \vspace{-3mm}
    \caption{Visualized results of top-5 physical knowledge relationship, where `top-5' indicates the five objects that have similar characteristics to the origin, and `physical property' indicates the similar physical properties of these objects.}
    \label{figure_top_k}
    \vspace{-2mm}
\end{figure*}

\subsubsection{CLM}
The Counterfactual Intervention Module (CLM) is designed to establish relationships based on physical knowledge across static, dynamic, and audio features, while also implementing counterfactual relation intervention to enhance reasoning capabilities and interpretability. To evaluate the effectiveness of the CLM, we analyze its performance from two perspectives: (1) the effectiveness of physical knowledge relationships and (2) the effectiveness of counterfactual relation intervention.

{\bf Effectiveness of physical knowledge relationships.} The physical knowledge relationship aims to aggregate features of objects with similar or identical physical properties. As shown in rows 2, 5, 9, and 10 of Table \ref{table:CLM}, establishing relationships among objects within a batch without decoupling does not improve performance. For instance, `CLIP A' experiences a 20.9\% decline on the PACS-Material subset, and `AudioCLIP A' decreases by 5.6\%. However, as indicated in rows 3 and 4 of Table \ref{table:CLM}, incorporating the DSE before applying the physical knowledge relationship improves accuracy on PACS by 1.5\%. Furthermore, with the proposed DSE+, accuracy increases to 3.5\%. Similar trends are observed in the PACS-Material subset. These results suggest that the Physical Knowledge Relationship cannot effectively model visual features mixed with extraneous information; instead, it requires purer features, such as decoupled static and dynamic features.

Moreover, Figure~\ref{figure_top_k} illustrates four objects and their top-5 similar objects. For example, the first row shows an object characterized by softness, with four of its top-5 similar objects also exhibiting softness. This demonstrates that the physical knowledge relationship successfully models the property of softness and aids in reasoning about objects with similar properties. Similarly, the third row features an object characterized by rockiness, and its top-5 similar objects predominantly share this characteristic. While the puzzle in the fourth column does not exhibit rockiness, we attribute this discrepancy to noise in the physical knowledge relationship, which is expected to diminish as the dataset size increases.

To further evaluate the effectiveness of the physical knowledge relationship, we tested different values of $K$. As shown in Figure~\ref{fig: T of DSE}(c) and (d), varying $K$ significantly impacts the results. Specifically, when $K = 2$ or $K = 8$, the accuracy on PACS slightly decreases, while the optimal performance is achieved at $K = 5$. This occurs because a small $K$ value introduces insufficient physical knowledge, whereas a large $K$ value introduces noise into the relationships. Notably, after incorporating the counterfactual module, the model's sensitivity to the $K$ value decreases, demonstrating that the counterfactual module enhances the reasoning capability of physical knowledge relationships.
 
{\bf Effectiveness of Counterfactual Relation Intervention.}~Table \ref{table:CLM} presents the results of an ablation study on Counterfactual Relation Intervention. As shown in rows 5 and 12, applying intervention `C' to relationship `A' improves accuracy on the PACS dataset by 1.9\% and 1.0\%, respectively. While the improvement is not significant compared to the CLIP baseline, the intervention successfully mitigates the negative impact associated with `A'. These findings further validate that the physical knowledge relationship can effectively model both static and dynamic factors, leading to more accurate relationship modeling after intervention. 

\begin{table*}[t]
\caption{The accuracy under incomplete modality conditions is evaluated for three scenarios: (1) ``Audio" is missing; (2) ``Video" is missing; and (3) ``Audio\&Video" both are missing simultaneously. ``N/A'' denotes results were unavailable.}
\vspace{-2mm}
\begin{tabularx}{\textwidth}{l *{7}{>{\centering\arraybackslash}X}}
\toprule
\multicolumn{1}{c}{Missing data / Method} & High Boundary & 10\%  & 30\%  & 50\%  & 70\%  & 90\%  & Low Boundary \\ \hline        
Audio / DCL                         & 65.3         & 61.7 & 61.2 & 60.6 & 54.1 & 58.5  & 59.0        \\
Audio / RDCL                          &  65.3         & {\bf 64.1} & {\bf 63.2} & {\bf 62.6} & {\bf 60.5} & {\bf 61.8} & 59.0        \\ \hline
Video / DCL                         & 65.3         & 62.4 & 60.5 & 60.0 & 60.1 & 59.3  & 58.1        \\
Video / RDCL                      & 65.3         & {\bf 63.1} & {\bf 63.1} & {\bf 62.5} & {\bf 61.4} & {\bf 61.1} & 58.1        \\ \hline
Audio \& Video / DCL                 & 65.3         & 61.3 & 58.8  & N/A  & N/A     & N/A     & 50.4         \\
Audio \& Video / RDCL             & 65.3         & {\bf 62.5}  & {\bf 59.6}  & N/A     & N/A     & N/A     & 50.4         \\ \bottomrule
\end{tabularx}
\label{incomplete modalities}
\end{table*}

\begin{table}[t]
\centering
\footnotesize 
\caption{Performance evaluation of different modal inputs on AudioCLIP~\cite{guzhov2022audioclip}. (I: Image, V: Video, A: Audio, T: VLM-Assisted Reasoning). ``$\checkmark$'' indicates the presence of a specific input modality.}
\vspace{-2mm}
\begin{adjustbox}{width=0.45\textwidth} 
\begin{tabular}{ccccc@{\hskip 10pt}c@{\hskip 12pt}c}
\toprule
I & A & V & T & DCL & \multicolumn{2}{c}{Accuracy (\%)} \\
 & & & & & PACS & PACS-Material \\ \midrule
$\checkmark$ &   &   &   &   & 59.2  & 73.5       \\
& $\checkmark$ &   &   &   & 57.9  & 66.0       \\
&   & $\checkmark$ &   &   & 58.7  & 70.2       \\
&   &   & $\checkmark$ &   & 63.4      & 77.0           \\
$\checkmark$ & $\checkmark$ & $\checkmark$ &   &   & 60.0  & 75.9       \\
$\checkmark$ &   &   &   & $\checkmark$ & 60.1 & 76.3 \\
& $\checkmark$ &   &   & $\checkmark$ & 58.2  & 69.4       \\
&   & $\checkmark$ &   & $\checkmark$ & \underline{64.4}   &  \underline{77.4}   \\
$\checkmark$   &  $\checkmark$ &  $\checkmark$ & $\checkmark$ & $\checkmark$ & {\bf 66.5}  & {\bf 80.3}       \\
\bottomrule
\end{tabular}
\end{adjustbox}
\label{table: Visual Bias}
\end{table}

\subsection{The Results of Incomplete Modalities}
In this section, we evaluate the effectiveness of the Incomplete Multi-Modal Learning Module (IMLM) introduced in Section~\ref{Incomplete modality learning Module}. First, we define the lower-bound and upper-bound metrics and describe the dataset composition. Subsequently, we compare the performance of various methods on datasets with incomplete modalities.

\subsubsection{Lower-Bound and Upper-Bound of the results}
Following~\cite{ma2021smil}, we establish the following experimental scenarios:
\begin{itemize}
\item \textbf{Lower-Bound} involves training using only a single modality, such as exclusively using single 100\% video data or 100\% audio data. These results represent the baseline performance for single-modal learning.
\item \textbf{Upper-Bound}: This scenario involves training using two complete modalities simultaneously. In our experiments, we use 100\% video data and 100\% audio data to establish the upper-bound performance.
\item \textbf{Missing ratio $\alpha$ of Data}: It simulates incomplete data conditions. When video data is missing, we use 100\% audio data combined with $\alpha_v$ video data. Conversely, when audio data is missing, we use 100\% video data combined with $\alpha_a$ audio data.
\end{itemize}

\subsubsection{Experiment Results}
As illustrated in Table~\ref{incomplete modalities}, we employ AudioCLIP as the baseline to demonstrate the performance of DCL and RDCL under conditions of modality incompleteness. In the first and second rows of Table~\ref{incomplete modalities}, the accuracy on PACS progressively declines as the rate of missing data $\alpha_a$ increases (from 61.5 at $\alpha_a=10\%$ to 54.1 at $\alpha_a=70\%$). Our proposed RDCL for modality data completion can mitigate this decline in accuracy (from 61.7 to 64.1 at $\alpha_a=10\%$, approaching the High Boundary 65.3). A similar trend can be observed in the results w.r.t video data, as shown in the 3-th and 4-th rows. A comparison of rows 2 and 4 shows that RDCL performs better under missing-audio than missing-video, as it leverages both static and dynamic factors to supplement shared features, whereas only audio features are available for missing-audio. When both modalities are missing (rows 5 and 6), RDCL effectively extracts shared semantic information representing physical knowledge from available features to compensate for the missing data, highlighting its robust resilience.

\section{Analysis}

\subsection{Impact of visual bias}
\label{Impact of visual bias}

\begin{figure}
    \centering
    \includegraphics[width=1.0\linewidth]{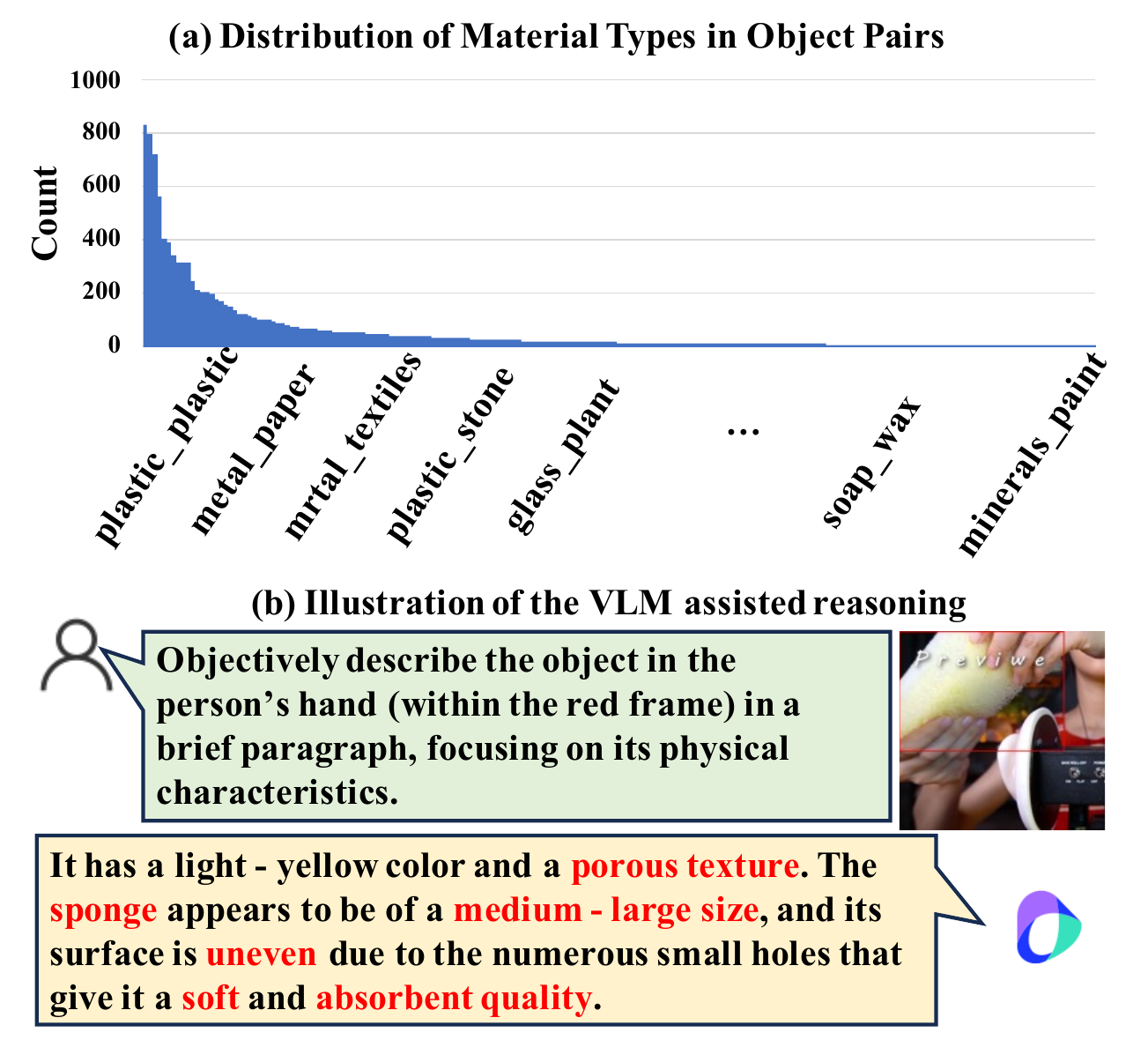}
    \caption{(a) The material type distribution of object pairs in the training set. (b) Example of the prompt and VLM generated responses.}
    \label{fig: fig_6}
\end{figure}

Visual information strongly biases model predictions due to its frequent co-occurrence with specific labels. As shown in Fig~\ref{fig: fig_6}(a), we examined material types for object pairs in the PACS dataset and observed a long-tail distribution, with some combinations appearing far more frequently than others. For example, the <plastic, metal> pair appears 370 times more frequently than the <plastic, styrofoam> pair, which occurs only 3 times. This imbalance causes models to depend on visual features, leading to incorrect predictions overly. When encountering the rare <plastic, styrofoam> pair, the model might mistakenly classify the second object as “metal” due to learned visual biases.
Table~\ref{table: Visual Bias} shows that the model underperformed in single-modality setups (I, V, A) compared to the full multimodal approach (I+A+V). By integrating our proposed Decoupled Contrastive Learning (DCL) framework into the video modality (V w/ DCL), the model successfully disentangled visual information, built physical knowledge relationships, reduced visual bias, and achieved improved results. Further analysis and results are available in the supplementary materials. 

\subsection{Impact of VLM-Assisted Reasoning}
\label{Impact of material information}
To further leverage the reasoning ability of the large language-vision model, we employed a widely used Vision-Language Model (Doubao-1.5-vision-pro) to generate descriptive interpretations of the input visual information for its physical properties, as shown in Fig~\ref{fig: fig_6}(b). These descriptions were then incorporated as additional inputs into our proposed model. As illustrated in Table~\ref{table: Visual Bias}, the inclusion of VLM-generated auxiliary reasoning information (I, A, V, T) further enhanced the model’s performance compared to using only visual and auditory data (I, A, V). This improvement demonstrates the utility of VLM-derived insights in aiding model inference. More details are provided in the supplementary materials.


\section{Conclusion}
In this paper, we presented a Robust Disentangled Counterfactual Learning for physical audiovisual commonsense reasoning, in which a Disentangled Sequential Encoder decoupled the video into time-invariant and time-varied factors, respectively. Furthermore, we modeled the physical knowledge relationship among objects as an affinity matrix and apply counterfactual relation intervention to emphasize the physical commonalities. In addition, an incomplete multi-modal learning method was utilized to recover the missing modality and alleviate the noisy disruption. As a plug-and-play component, our method can be readily incorporated and experimental results demonstrated its potential to significantly enhance multiple baselines. In the future, we will apply our proposed method to robotic and embodied AI.

\bibliographystyle{IEEEtran}
\bibliography{reference}

\begin{thebibliography}{10}
\providecommand{\url}[1]{#1}
\csname url@samestyle\endcsname
\providecommand{\newblock}{\relax}
\providecommand{\bibinfo}[2]{#2}
\providecommand{\BIBentrySTDinterwordspacing}{\spaceskip=0pt\relax}
\providecommand{\BIBentryALTinterwordstretchfactor}{4}
\providecommand{\BIBentryALTinterwordspacing}{\spaceskip=\fontdimen2\font plus
\BIBentryALTinterwordstretchfactor\fontdimen3\font minus \fontdimen4\font\relax}
\providecommand{\BIBforeignlanguage}[2]{{%
\expandafter\ifx\csname l@#1\endcsname\relax
\typeout{** WARNING: IEEEtran.bst: No hyphenation pattern has been}%
\typeout{** loaded for the language `#1'. Using the pattern for}%
\typeout{** the default language instead.}%
\else
\language=\csname l@#1\endcsname
\fi
#2}}
\providecommand{\BIBdecl}{\relax}
\BIBdecl

\bibitem{kriegeskorte2015deep}
N.~Kriegeskorte, ``Deep neural networks: a new framework for modeling biological vision and brain information processing,'' \emph{Annual review of vision science}, vol.~1, pp. 417--446, 2015.

\bibitem{yu2022pacs}
S.~Yu, P.~Wu, P.~P. Liang, R.~Salakhutdinov, and L.-P. Morency, ``Pacs: A dataset for physical audiovisual commonsense reasoning,'' in \emph{Proc. Eur. Conf. Comput. Vis}, 2022, pp. 292--309.

\bibitem{an2024etpnav}
D.~An, H.~Wang, W.~Wang, Z.~Wang, Y.~Huang, K.~He, and L.~Wang, ``Etpnav: Evolving topological planning for vision-language navigation in continuous environments,'' \emph{IEEE Trans. Pattern Anal. Mach. Intell.}, 2024.

\bibitem{purushwalkam2021audio}
S.~Purushwalkam, S.~V.~A. Gari, V.~K. Ithapu, C.~Schissler, P.~Robinson, A.~Gupta, and K.~Grauman, ``Audio-visual floorplan reconstruction,'' in \emph{Proc. IEEE Int. Conf. Comput. Vis.}, 2021, pp. 1183--1192.

\bibitem{chen2021semantic}
C.~Chen, Z.~Al-Halah, and K.~Grauman, ``Semantic audio-visual navigation,'' in \emph{Proc. IEEE Conf. Comput. Vis. Pattern Recognit.}, 2021, pp. 15\,516--15\,525.

\bibitem{hou2019deep}
M.~Hou, J.~Tang, J.~Zhang, W.~Kong, and Q.~Zhao, ``Deep multimodal multilinear fusion with high-order polynomial pooling,'' \emph{Proc. Adv. Neural Inf. Process. Syst.}, vol.~32, 2019.

\bibitem{peng2022balanced}
X.~Peng, Y.~Wei, A.~Deng, D.~Wang, and D.~Hu, ``Balanced multimodal learning via on-the-fly gradient modulation,'' in \emph{Proc. IEEE Conf. Comput. Vis. Pattern Recognit.}, 2022, pp. 8238--8247.

\bibitem{lv2024disentangled}
C.~Lv, S.~Zhang, Y.~Tian, M.~Qi, and H.~Ma, ``Disentangled counterfactual learning for physical audiovisual commonsense reasoning,'' \emph{Proc. Adv. Neural Inf. Process. Syst.}, vol.~36, 2023.

\bibitem{piloto2022intuitive}
L.~S. Piloto, A.~Weinstein, P.~Battaglia, and M.~Botvinick, ``Intuitive physics learning in a deep-learning model inspired by developmental psychology,'' \emph{Nature human behaviour}, vol.~6, no.~9, pp. 1257--1267, 2022.

\bibitem{hespos2016five}
S.~J. Hespos, A.~L. Ferry, E.~M. Anderson, E.~N. Hollenbeck, and L.~J. Rips, ``Five-month-old infants have general knowledge of how nonsolid substances behave and interact,'' \emph{Psychological Science}, vol.~27, no.~2, pp. 244--256, 2016.

\bibitem{zellers2019recognition}
R.~Zellers, Y.~Bisk, A.~Farhadi, and Y.~Choi, ``From recognition to cognition: Visual commonsense reasoning,'' in \emph{Proc. IEEE Conf. Comput. Vis. Pattern Recognit.}, 2019, pp. 6720--6731.

\bibitem{li2022representation}
J.~Li, L.~Niu, and L.~Zhang, ``From representation to reasoning: Towards both evidence and commonsense reasoning for video question-answering,'' in \emph{Proc. IEEE Conf. Comput. Vis. Pattern Recognit.}, 2022, pp. 21\,273--21\,282.

\bibitem{bisk2020piqa}
Y.~Bisk, R.~Zellers, J.~Gao, Y.~Choi \emph{et~al.}, ``Piqa: Reasoning about physical commonsense in natural language,'' in \emph{AAAI Conference on Artificial Intelligence}, vol.~34, no.~05, 2020, pp. 7432--7439.

\bibitem{zellers2022merlot}
R.~Zellers, J.~Lu, X.~Lu, Y.~Yu, Y.~Zhao, M.~Salehi, A.~Kusupati, J.~Hessel, A.~Farhadi, and Y.~Choi, ``Merlot reserve: Neural script knowledge through vision and language and sound,'' in \emph{Proc. IEEE Conf. Comput. Vis. Pattern Recognit.}, 2022, pp. 16\,375--16\,387.

\bibitem{wang2020visual}
T.~Wang, J.~Huang, H.~Zhang, and Q.~Sun, ``Visual commonsense r-cnn,'' in \emph{Proc. IEEE Conf. Comput. Vis. Pattern Recognit.}, 2020, pp. 10\,760--10\,770.

\bibitem{zareian2020learning}
A.~Zareian, Z.~Wang, H.~You, and S.-F. Chang, ``Learning visual commonsense for robust scene graph generation,'' in \emph{Proc. Eur. Conf. Comput. Vis.}, 2020, pp. 642--657.

\bibitem{lin2023tiktalk}
H.~Lin, L.~Ruan, W.~Xia, P.~Liu, J.~Wen, Y.~Xu, D.~Hu, R.~Song, W.~X. Zhao, Q.~Jin \emph{et~al.}, ``Tiktalk: A video-based dialogue dataset for multi-modal chitchat in real world,'' in \emph{Proc. ACM Int. Conf. on Multimedia}, 2023, pp. 1303--1313.

\bibitem{bengio2013representation}
Y.~Bengio, A.~Courville, and P.~Vincent, ``Representation learning: A review and new perspectives,'' \emph{IEEE Trans. Pattern Anal. Mach. Intell.}, vol.~35, no.~8, pp. 1798--1828, 2013.

\bibitem{chen2021curriculum}
H.~Chen, Y.~Chen, X.~Wang, R.~Xie, R.~Wang, F.~Xia, and W.~Zhu, ``Curriculum disentangled recommendation with noisy multi-feedback,'' \emph{Proc. Adv. Neural Inf. Process. Syst.}, vol.~34, pp. 26\,924--26\,936, 2021.

\bibitem{wei2024unsupervised}
P.~Wei, L.~Kong, X.~Qu, Y.~Ren, Z.~Xu, J.~Jiang, and X.~Yin, ``Unsupervised video domain adaptation for action recognition: A disentanglement perspective,'' \emph{Proc. Adv. Neural Inf. Process. Syst.}, vol.~36, 2024.

\bibitem{qi2020stc}
M.~Qi, Y.~Wang, A.~Li, and J.~Luo, ``Stc-gan: Spatio-temporally coupled generative adversarial networks for predictive scene parsing,'' \emph{IEEE Trans. Image Process.}, vol.~29, pp. 5420--5430, 2020.

\bibitem{van2019disentangled}
S.~Van~Steenkiste, F.~Locatello, J.~Schmidhuber, and O.~Bachem, ``Are disentangled representations helpful for abstract visual reasoning?'' \emph{Proc. Adv. Neural Inf. Process. Syst.}, vol.~32, 2019.

\bibitem{qi2021semantics}
M.~Qi, J.~Qin, Y.~Yang, Y.~Wang, and J.~Luo, ``Semantics-aware spatial-temporal binaries for cross-modal video retrieval,'' \emph{IEEE Trans. Image Process.}, vol.~30, pp. 2989--3004, 2021.

\bibitem{ma2018disentangled}
L.~Ma, Q.~Sun, S.~Georgoulis, L.~Van~Gool, B.~Schiele, and M.~Fritz, ``Disentangled person image generation,'' in \emph{Proc. IEEE Conf. Comput. Vis. Pattern Recognit.}, 2018, pp. 99--108.

\bibitem{bai2021contrastively}
J.~Bai, W.~Wang, and C.~P. Gomes, ``Contrastively disentangled sequential variational autoencoder,'' \emph{Proc. Adv. Neural Inf. Process. Syst.}, vol.~34, pp. 10\,105--10\,118, 2021.

\bibitem{wang2024rdfc}
H.~Wang, Z.~Che, Y.~Yang, M.~Wang, Z.~Xu, X.~Qiao, M.~Qi, F.~Feng, and J.~Tang, ``Rdfc-gan: Rgb-depth fusion cyclegan for indoor depth completion,'' \emph{IEEE Trans. Pattern Anal. Mach. Intell.}, 2024.

\bibitem{qi2019attentive}
M.~Qi, W.~Li, Z.~Yang, Y.~Wang, and J.~Luo, ``Attentive relational networks for mapping images to scene graphs,'' in \emph{Proc. IEEE Conf. Comput. Vis. Pattern Recognit.}, 2019, pp. 3957--3966.

\bibitem{qi2019ke}
M.~Qi, Y.~Wang, J.~Qin, and A.~Li, ``Ke-gan: Knowledge embedded generative adversarial networks for semi-supervised scene parsing,'' in \emph{Proc. IEEE Conf. Comput. Vis. Pattern Recognit.}, 2019, pp. 5237--5246.

\bibitem{tran2017disentangled}
L.~Tran, X.~Yin, and X.~Liu, ``Disentangled representation learning gan for pose-invariant face recognition,'' in \emph{Proc. IEEE Conf. Comput. Vis. Pattern Recognit.}, 2017, pp. 1415--1424.

\bibitem{goodfellow2020generative}
I.~Goodfellow, J.~Pouget-Abadie, M.~Mirza, B.~Xu, D.~Warde-Farley, S.~Ozair, A.~Courville, and Y.~Bengio, ``Generative adversarial networks,'' \emph{Communications of the ACM}, vol.~63, no.~11, pp. 139--144, 2020.

\bibitem{zhu2020s3vae}
Y.~Zhu, M.~R. Min, A.~Kadav, and H.~P. Graf, ``S3vae: Self-supervised sequential vae for representation disentanglement and data generation,'' in \emph{Proc. IEEE Conf. Comput. Vis. Pattern Recognit.}, 2020, pp. 6538--6547.

\bibitem{wang2023disavr}
Y.~Wang, B.~Wei, J.~Liu, L.~Zhang, J.~Wang, and Q.~Wang, ``Disavr: Disentangled adaptive visual reasoning network for diagram question answering,'' \emph{IEEE Trans. Image Process.}, 2023.

\bibitem{goyal2017making}
Y.~Goyal, T.~Khot, D.~Summers-Stay, D.~Batra, and D.~Parikh, ``Making the v in vqa matter: Elevating the role of image understanding in visual question answering,'' in \emph{Proc. IEEE Conf. Comput. Vis. Pattern Recognit.}, 2017, pp. 6904--6913.

\bibitem{antol2015vqa}
S.~Antol, A.~Agrawal, J.~Lu, M.~Mitchell, D.~Batra, C.~L. Zitnick, and D.~Parikh, ``Vqa: Visual question answering,'' in \emph{Proc. IEEE Int. Conf. Comput. Vis.}, 2015, pp. 2425--2433.

\bibitem{goyal2019counterfactual}
Y.~Goyal, Z.~Wu, J.~Ernst, D.~Batra, D.~Parikh, and S.~Lee, ``Counterfactual visual explanations,'' in \emph{Proc. Inter. Conf. on Mach. Learn.}, 2019, pp. 2376--2384.

\bibitem{tang2020unbiased}
K.~Tang, Y.~Niu, J.~Huang, J.~Shi, and H.~Zhang, ``Unbiased scene graph generation from biased training,'' in \emph{Proc. IEEE Conf. Comput. Vis. Pattern Recognit.}, 2020, pp. 3716--3725.

\bibitem{Rao_2021_ICCV}
Y.~Rao, G.~Chen, J.~Lu, and J.~Zhou, ``Counterfactual attention learning for fine-grained visual categorization and re-identification,'' in \emph{Proc. IEEE Int. Conf. Comput. Vis.}, 2021, pp. 1025--1034.

\bibitem{xu2022unintentional}
J.~Xu, G.~Chen, J.~Lu, and J.~Zhou, ``Unintentional action localization via counterfactual examples,'' \emph{IEEE Trans. Image Process.}, vol.~31, pp. 3281--3294, 2022.

\bibitem{sun2023unbiased}
S.~Sun, S.~Zhi, Q.~Liao, J.~Heikkil{\"a}, and L.~Liu, ``Unbiased scene graph generation via two-stage causal modeling,'' \emph{IEEE Trans. Pattern Anal. Mach. Intell.}, vol.~45, no.~10, pp. 12\,562--12\,580, 2023.

\bibitem{xue2023variational}
D.~Xue, S.~Qian, and C.~Xu, ``Variational causal inference network for explanatory visual question answering,'' in \emph{Proc. IEEE Int. Conf. Comput. Vis.}, 2023, pp. 2515--2525.

\bibitem{li2023progressive}
G.~Li, W.~Hou, and D.~Hu, ``Progressive spatio-temporal perception for audio-visual question answering,'' in \emph{Proceedings of the 31st ACM International Conference on Multimedia}, 2023, pp. 7808--7816.

\bibitem{lu2019vilbert}
J.~Lu, D.~Batra, D.~Parikh, and S.~Lee, ``Vilbert: Pretraining task-agnostic visiolinguistic representations for vision-and-language tasks,'' \emph{Proc. Adv. Neural Inf. Process. Syst.}, vol.~32, 2019.

\bibitem{hu2021class}
D.~Hu, Y.~Wei, R.~Qian, W.~Lin, R.~Song, and J.-R. Wen, ``Class-aware sounding objects localization via audiovisual correspondence,'' \emph{IEEE Trans. Pattern Anal. Mach. Intell.}, 2021.

\bibitem{deshmukh2023pengi}
S.~Deshmukh, B.~Elizalde, R.~Singh, and H.~Wang, ``Pengi: An audio language model for audio tasks,'' \emph{Proc. Adv. Neural Inf. Process. Syst.}, vol.~36, pp. 18\,090--18\,108, 2023.

\bibitem{li2022learning}
G.~Li, Y.~Wei, Y.~Tian, C.~Xu, J.-R. Wen, and D.~Hu, ``Learning to answer questions in dynamic audio-visual scenarios,'' in \emph{Proc. IEEE Conf. Comput. Vis. Pattern Recognit.}, 2022, pp. 19\,108--19\,118.

\bibitem{wang2023multi}
H.~Wang, Y.~Chen, C.~Ma, J.~Avery, L.~Hull, and G.~Carneiro, ``Multi-modal learning with missing modality via shared-specific feature modelling,'' in \emph{Proc. IEEE Conf. Comput. Vis. Pattern Recognit.}, 2023, pp. 15\,878--15\,887.

\bibitem{graves2005framewise}
A.~Graves and J.~Schmidhuber, ``Framewise phoneme classification with bidirectional lstm and other neural network architectures,'' \emph{Neural networks}, vol.~18, no. 5-6, pp. 602--610, 2005.

\bibitem{kingma2013auto}
D.~P. Kingma and M.~Welling, ``Auto-encoding variational bayes,'' \emph{arXiv preprint arXiv:1312.6114}, 2013.

\bibitem{khosla2020supervised}
P.~Khosla, P.~Teterwak, C.~Wang, A.~Sarna, Y.~Tian, P.~Isola, A.~Maschinot, C.~Liu, and D.~Krishnan, ``Supervised contrastive learning,'' \emph{Proc. Adv. Neural Inf. Process. Syst.}, vol.~33, pp. 18\,661--18\,673, 2020.

\bibitem{chen2020simple}
T.~Chen, S.~Kornblith, M.~Norouzi, and G.~Hinton, ``A simple framework for contrastive learning of visual representations,'' in \emph{Proc. Inter. Conf. on Mach. Learn.}, 2020, pp. 1597--1607.

\bibitem{ding2022mukea}
Y.~Ding, J.~Yu, B.~Liu, Y.~Hu, M.~Cui, and Q.~Wu, ``Mukea: Multimodal knowledge extraction and accumulation for knowledge-based visual question answering,'' in \emph{Proc. IEEE Conf. Comput. Vis. Pattern Recognit.}, 2022, pp. 5089--5098.

\bibitem{pearl2018book}
J.~Pearl and D.~Mackenzie, \emph{The book of why: the new science of cause and effect}.\hskip 1em plus 0.5em minus 0.4em\relax Basic books, 2018.

\bibitem{pandeya2021deep}
Y.~R. Pandeya and J.~Lee, ``Deep learning-based late fusion of multimodal information for emotion classification of music video,'' \emph{Multimedia Tools and Applications}, vol.~80, pp. 2887--2905, 2021.

\bibitem{he2016deep}
K.~He, X.~Zhang, S.~Ren, and J.~Sun, ``Deep residual learning for image recognition,'' in \emph{Proc. IEEE Conf. Comput. Vis. Pattern Recognit.}, 2016, pp. 770--778.

\bibitem{radford2021learning}
A.~Radford, J.~W. Kim, C.~Hallacy, A.~Ramesh, G.~Goh, S.~Agarwal, G.~Sastry, A.~Askell, P.~Mishkin, J.~Clark \emph{et~al.}, ``Learning transferable visual models from natural language supervision,'' in \emph{Proc. Inter. Conf. on Mach. Learn.}, 2021, pp. 8748--8763.

\bibitem{guzhov2022audioclip}
A.~Guzhov, F.~Raue, J.~Hees, and A.~Dengel, ``Audioclip: Extending clip to image, text and audio,'' in \emph{ICASSP 2022-2022 IEEE International Conference on Acoustics, Speech and Signal Processing (ICASSP)}.\hskip 1em plus 0.5em minus 0.4em\relax IEEE, 2022, pp. 976--980.

\bibitem{chen2020uniter}
Y.-C. Chen, L.~Li, L.~Yu, A.~El~Kholy, F.~Ahmed, Z.~Gan, Y.~Cheng, and J.~Liu, ``Uniter: Universal image-text representation learning,'' in \emph{Proc. Eur. Conf. Comput. Vis.}, 2020, pp. 104--120.

\bibitem{suhr2019corpus}
A.~Suhr, S.~Zhou, A.~Zhang, I.~Zhang, H.~Bai, and Y.~Artzi, ``A corpus for reasoning about natural language grounded in photographs,'' in \emph{Proceedings of the 57th Annual Meeting of the Association for Computational Linguistics}, 2019, pp. 6418--6428.

\bibitem{team2024gemini}
G.~Team, P.~Georgiev, V.~I. Lei, R.~Burnell, L.~Bai, A.~Gulati, G.~Tanzer, D.~Vincent, Z.~Pan, S.~Wang \emph{et~al.}, ``Gemini 1.5: Unlocking multimodal understanding across millions of tokens of context,'' \emph{arXiv preprint arXiv:2403.05530}, 2024.

\bibitem{achiam2023gpt}
J.~Achiam, S.~Adler, S.~Agarwal, L.~Ahmad, I.~Akkaya, F.~L. Aleman, D.~Almeida, J.~Altenschmidt, S.~Altman, S.~Anadkat \emph{et~al.}, ``Gpt-4 technical report,'' \emph{arXiv preprint arXiv:2303.08774}, 2023.

\bibitem{bai2023qwen}
J.~Bai, S.~Bai, S.~Yang, S.~Wang, S.~Tan, P.~Wang, J.~Lin, C.~Zhou, and J.~Zhou, ``Qwen-vl: A frontier large vision-language model with versatile abilities,'' \emph{arXiv preprint arXiv:2308.12966}, 2023.

\bibitem{ma2021smil}
M.~Ma, J.~Ren, L.~Zhao, S.~Tulyakov, C.~Wu, and X.~Peng, ``Smil: Multimodal learning with severely missing modality,'' in \emph{AAAI Conference on Artificial Intelligence}, vol.~35, no.~3, 2021, pp. 2302--2310.

\end{thebibliography}




\end{document}


\renewcommand\arraystretch{1.5}
%
\title{\title{Supplementary Materials of  \\ \emph{Robust Disentangled Counterfactual Learning for Physical Audiovisual Commonsense Reasoning}}}
%
%
%
%

\author{
        Mengshi Qi,~\IEEEmembership{Member,~IEEE},
        Changsheng Lv,
        Huadong Ma,~\IEEEmembership{Fellow,~IEEE}
\thanks{This work is partly supported by the Funds for the NSFC Project under Grant 62202063, Beijing Natural Science Foundation (L243027). (\emph{Corresponding author: Mengshi Qi~(email:~qms@bupt.edu.cn)})}
\thanks{M. Qi, C. Lv, and H. Ma are with the State Key Laboratory of Networking and Switching Technology, Beijing University of Posts and Telecommunications, China.}
}
%
%

\markboth{Transactions on Pattern Analysis and Machine Intelligence}%
{Shell \MakeLowercase{\textit{et al.}}: Bare Demo of IEEEtran.cls for Computer Society Journals}
%



\IEEEtitleabstractindextext{%

\begin{IEEEkeywords}
Physical Commonsense Reasoning, Robust Multimodal Learning, Disentangled Representation, Counterfactual Analysis.
\end{IEEEkeywords}}

\maketitle

\IEEEdisplaynontitleabstractindextext

%
\IEEEpeerreviewmaketitle

In this supplementary material, we provide a comprehensive algorithm underlying our proposed model, encompassing both the DCL and RDCL in Section~\ref{Sec. Algorithm of DCL and RDCL}. Section~\ref{Sec: Drivations and more experimental results} includes derivations and supplementary experimental results. Additionally, Section~\ref{Sec: VLM-Assisted Reasoning Dataset} presents more samples and statistical analyses of the VLM-Assisted Reasoning Dataset introduced in our main paper.

%
%
%
%
\section{Algorithm of DCL and RDCL}
In this section, we introduce the detailed algorithms for Disentangled Counterfactual
Learning (DCL) in Section~\ref{Sec. Algorithm of DCL} and Robust Disentangled Counterfactual
Learning (RDCL) in Section~\ref{Sec. Algorithm of RDCL} for Physical Commonsense Reasoning.  
\label{Sec. Algorithm of DCL and RDCL}
\subsection{DCL}
\label{Sec. Algorithm of DCL}
The overall framework of the proposed DCL algorithm is outlined in Algorithm~\ref{algorithm: DCL}. The model takes as input a training batch consisting of paired video-audio data along with associated physical knowledge questions. It outputs the final prediction, denoted as $\hat{Y}_{TIE}$.  

\subsection{RDCL}
\label{Sec. Algorithm of RDCL}
Unlike DCL, which processes complete multimodal inputs, RDCL is designed to handle incomplete modalities. As an illustrative example, we consider scenarios where audio data are missing. The corresponding algorithm is presented in Algorithm~\ref{algorithm: RDCL}.
\begin{algorithm}[ht]
\caption{Disentangled Counterfactual Learning~(DCL) Batch-Wise Training}
\label{algorithm: DCL}
\SetAlgoLined
\KwIn{
    Training batch $\{ \langle v_1, v_2 \rangle_i, \langle a_1, a_2 \rangle_i, q_i \}_{i=1}^{B}$, \\
    Batch size $B$, \\
    Pretrained image encoder $\mathcal{E}_{\text{img}}(\theta)$, \\
    Pretrained audio encoder $\mathcal{E}_{\text{aud}}(\theta)$, \\
    Pretrained text encoder $\mathcal{E}_{\text{text}}(\theta)$, \\
    Labels $\{Y_{GT,i}\}_{i=1}^B$, \\
    Number of frames $T$
}
\KwOut{
    Predicted answers $\{\hat{Y}_{TIE,i}\}_{i=1}^B$
}

Encode features: \\
  \For{$j \in \{1, 2\}$}{ 
   \quad $X^{v_j} = \{X_1^{v_j}, X_2^{v_j}, \cdots, X_T^{v_j}\} \gets \mathcal{E}_{\text{img}}(v_j)$ \\ 
   \quad $X^{a_j} \gets \mathcal{E}_{\text{aud}}(a_j)$
   }
   $X^t \gets \mathcal{E}_{\text{text}}(q)$ \\

\For{each sample in the batch}{
    Disentangle static factors $X^v_s$ and dynamic factors $X^v_z$ from $X^v$ via DSE in Section 4.2. \\
}

Compute the adjacency matrix $A_X$ using Eq. (15), (16), and (17). \\

Obtain the fused feature $F_1$, $F_2$ using Eq.(14). \\

Construct intervened features $X^*$ using Eq. (20), and compute the intervened adjacency matrix ${A}^*$ using Eqs. (15), (16) and (17). \\

Predict the $\hat{Y}_{X, A_X}$ and $\hat{Y}_{X^*, A_{X^*}}$ using Eq.(18)\\

Use $\hat{Y}_{TIE}$ obtained from Eq.(19) as the output.
\end{algorithm}

\section{Drivations and more experimental results}
\label{Sec: Drivations and more experimental results}
\subsection{Approximate Estimation of the Objective Function}

In Section 4.2 Disentangled Sequential Encoder of our main paper, our goal is to maximize the log-likelihood of \( x_{1:T} \). However, due to the computational complexity associated with high-dimensional integrals, directly obtaining \( \log p(x_{1:T}) \) is challenging. To address this issue, we employ the Evidence Lower Bound (ELBO) as an approximation to the log-likelihood. 

{\bf For the input sequence $x_{1:T}$}, Eq.(~\ref{eq:elbo_derivation}) in Figure~\ref{Seq-ELBO} shown adapted from the standard VAE framework \cite{bai2021contrastively}, noticing that either the prior or the approximate posterior factorizes over $s$ and $z_{1:T}$.
{\bf For the entire dataset}, let \( p_D \) represent the empirical data distribution, which assigns a probability mass of \( 1/N \) to each of the \( N \) training sequences in \( D \). The aggregated posteriors are defined as shown in Eq.(~\ref{Eq.2}),  Eq.(~\ref{Eq.3}), and  Eq.(~\ref{Eq.4}) in Figure~\ref{proof_3}. By rearranging terms and applying similar operations to \( x \), we arrive at Eq.~(\ref {Eq.6}) and Eq.~(\ref{Eq.7}) in Figure~\ref{proof_3}. Finally, integrating the above derivations, we obtain the dataset ELBO by subtracting a distinct KL divergence from the data log-likelihood, as illustrated in Eq.~\ref{eq:elbo} in Figure~\ref{proof-4}.
\begin{figure*}
\begin{equation}
\begin{aligned}
& \log p(x_{1:T}) \\
\ge& -KL[q(s,z_{1:T}|x_{1:T})||p(s,z_{1:T}|x_{1:T})]+\log p(x_{1:T})\\
=&\mathbb E_{q(s, z_{1:T}|x_{1:T})} \left[ \log p(s,z_{1:T}|x_{1:T}) - \log q(s,z_{1:T}|x_{1:T}) + \log p(x_{1:T}) \right] \\
=&\mathbb E_{q(s,z_{1:T}|x_{1:T})}[\log p(x_{1:T}|s, z_{1:T})-\log q(s,z_{1:T}|x_{1:T})+\log p(s,z_{1:T})]\\
=&\mathbb E_{q(s,z_{1:T}|x_{1:T})}[\log p(x_{1:T}|s, z_{1:T})-\log q(s|x_{1:T}) - \log p(z_{1:T}|x_{1:T})+\log p(s) + \log p(z_{1:T})]\\
=&\mathbb E_{q(z_{1:T}, s|x_{1:T})} \left[ \underbrace{\log p(x_{1:T}|s, z_{1:T})}_{\text{Reconstruction term}}-\underbrace{KL[q(s|x_{1:T})||p(s)]}_{s\text{-regression}}-\underbrace{KL[q(z_{1:T}|x_{1:T})||p(z_{1:T})]}_{z\text{-regression}} \right].
\end{aligned}  
\label{eq:elbo_derivation}
\end{equation}
\caption{The ELBO derivation for the input sequence \( x_{1:T} \).}
\label{Seq-ELBO}
\end{figure*}

\begin{figure*}
\centering
\begin{align}
q(s) & = \mathbb E_{x_{1:T}\sim p_D} [q(s|x_{1:T})] = \frac{1}{N} \sum_{x_{1:T}\in D} q(s|x_{1:T}), \label{Eq.2} \\
q(z_{1:T}) & = \mathbb E_{x_{1:T}\sim p_D} [q(z_{1:T}|x_{1:T})] = \frac{1}{N} \sum_{x_{1:T}\in D} q(z_{1:T}|x_{1:T}), \label{Eq.3} \\
q(s,z_{1:T}) & = \mathbb E_{x_{1:T}\sim p_D} [q(s|x_{1:T}) q(z_{1:T}|x_{1:T})] = \frac{1}{N} \sum_{x_{1:T}\in D} q(s|x_{1:T}) q(z_{1:T}|x_{1:T}). \label{Eq.4}
\end{align}
\begin{equation}
\begin{aligned}
& \mathbb E_{x_{1:T}\sim p_D}[KL[q(s|x_{1:T})||p(s)]] \\
=& \mathbb E_{x_{1:T}\sim p_D}\mathbb E_{q(s|x_{1:T})}[\log q(s|x_{1:T}) - \log q(s) + \log q(s) - \log p(s)] \\
=& \mathbb E_{q(x_{1:T}, s)} \log \left[ \frac{q(s|x_{1:T})}{q(s)} \right] + \mathbb E_{q(x_{1:T}, s)} [\log q(s)-\log p(s)] \\
=& I_q (x_{1:T};s) + KL [q(s)||p(s)]. 
\end{aligned}
\label{}
\end{equation}
\begin{equation}
KL [q(s)||p(s)] = \mathbb E_{x_{1:T}\sim p_D}[KL[q(s|x_{1:T})||p(s)]] - I_q(x_{1:T};s). 
\label{Eq.6}
\end{equation}
\begin{gather}
KL[q(z_{1:T})||p(z_{1:T})] = \mathbb E_{x_{1:T}\sim p_D}[KL[q(z_{1:T}|x_{1:T})||p(z_{1:T})]] - I_q(x_{1:T};z_{1:T}). 
\label{Eq.7}
\end{gather}
\caption{Aggregated equations and their relationships.}
\label{proof_3}
\end{figure*}

\begin{figure*}
\begin{equation}
\begin{aligned} 
&\frac{1}{N} \sum_{x_{1:T}\in D} \log p(x_{1:T}) = \mathbb E_{x_{1:T}\sim p_D}[\log p(x_{1:T})] \\
\ge& \mathbb E_{x_{1:T}\sim p_D}[\log p(x_{1:T}) - KL[q(s, z_{1:T})||p(s, z_{1:T}|x_{1:T})]] \\
=& \mathbb E_{x_{1:T}\sim p_D}[\mathbb E_{q(s, z_{1:T}|x_{1:T})}[\log p(x_{1:T}) - \log q(s, z_{1:T}) + \log p(s, z_{1:T}|x_{1:T})]] \\
=& \mathbb E_{x_{1:T}\sim p_D}[\mathbb E_{q(s, z_{1:T}|x_{1:T})}[\log p(x_{1:T}) - \log q(s, z_{1:T}) + \log p(x_{1:T}|s, z_{1:T}) + \log p(s, z_{1:T}) - \log p(x_{1:T})]] \\
=& \mathbb E_{x_{1:T}\sim p_D}[\mathbb E_{q(s, z_{1:T}|x_{1:T})}[\log p(x_{1:T}|s, z_{1:T}) - \log q(s, z_{1:T}) + \log p(s, z_{1:T})]] \\
=& \mathbb E_{x_{1:T}\sim p_D}[\mathbb E_{q(s, z_{1:T}|x_{1:T})}[\log p(x_{1:T}|s, z_{1:T})]] - \mathbb E_{x_{1:T}\sim p_D}[\mathbb E_{q(s, z_{1:T}|x_{1:T})}[\log q(s, z_{1:T}) - \log p(s, z_{1:T})]] \\
=& \mathbb E_{x_{1:T}\sim p_D}[\mathbb E_{q(s, z_{1:T}|x_{1:T})}[\log p(x_{1:T}|s, z_{1:T})]] - KL[q(s, z_{1:T})||p(s, z_{1:T})] \\
=& \mathbb E_{x_{1:T}\sim p_D}[\mathbb E_{q(s, z_{1:T}|x_{1:T})}[\log p(x_{1:T}|s, z_{1:T})]] - I_q(s;z_{1:T}) - KL[q(s)||p(s)] - KL[q(z_{1:T})||p(z_{1:T})] \\
=& \mathbb E_{x_{1:T}\sim p_D}[\mathbb E_{q(s, z_{1:T}|x_{1:T})}[\log p(x_{1:T}|s, z_{1:T})]] \\
&\hspace{5em} - \mathbb E_{x_{1:T}\sim p_D}[KL[q(s|x_{1:T})||p(s)]] - \mathbb E_{x_{1:T}\sim p_D}[KL[q(z_{1:T}|x_{1:T})||p(z_{1:T})]] \\
&\hspace{5em} + I_q(s;x_{1:T}) + I_q(z_{1:T};x_{1:T}) - I_q(s;z_{1:T}).
\end{aligned}
\label{eq:elbo}
\end{equation}
\caption{Derivation of the ELBO for a dataset by subtracting a KL-divergence term from the data log-likelihood.}
\label{proof-4}
\end{figure*}

\subsection{Sensitivity Analysis of Parameters}  
\begin{figure*}[ht]
    \centering
    \includegraphics[width=1.0\linewidth]{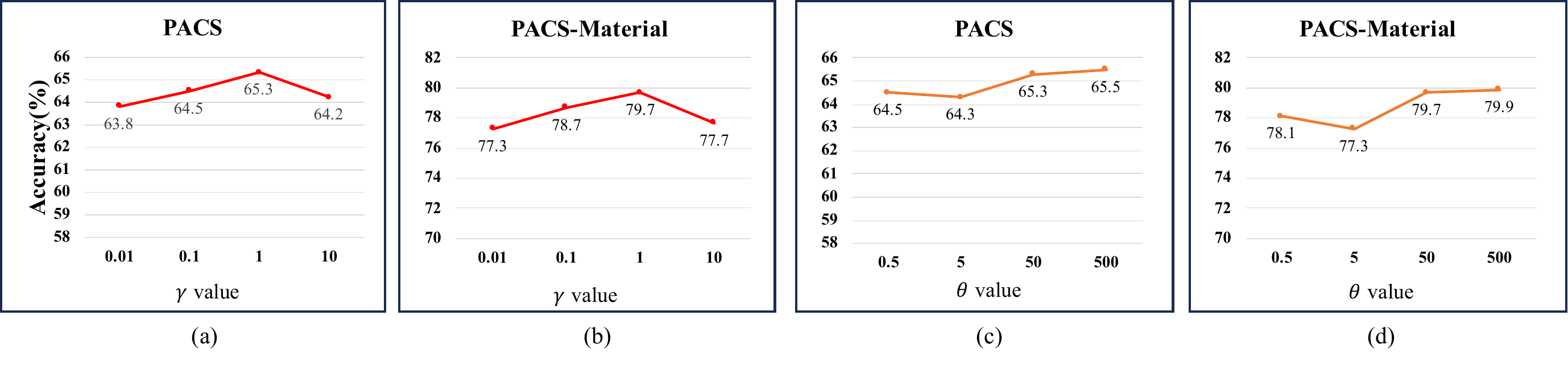}
    \caption{Performance comparisons of various hyperparameters in Eq. (6) are presented. Figures (a) and (b) display the performance of AudioCLIP with different values of $\gamma$ in $\mathcal{L}_{DSE}$ on the PACS and PACS-Material datasets. Figures (c) and (d) show the performance of AudioCLIP with varying $\theta$ in $\mathcal{L}_{DSE}$ on the same datasets.}
    \label{fig: hyperparameters}
\end{figure*}

We conducted a sensitivity analysis on the parameters $\gamma$ and $\theta$ as defined in Eq. (6) of the main paper, with results presented in Figure~\ref{fig: hyperparameters}. Specifically, we evaluated $\gamma$ over the range $\{0.01, 0.1, 1, 10\}$ and $\theta$ over the range $\{0.5, 5, 50, 500\}$. The results denote that our proposed DCL method exhibits strong robustness to variations in both $\gamma$ and $\theta$, achieving consistent and stable performance across all tested parameter configurations.
\subsection{Analysis of dynamic factors}
\begin{figure}[ht]
    \centering
    \includegraphics[width=0.9\linewidth]{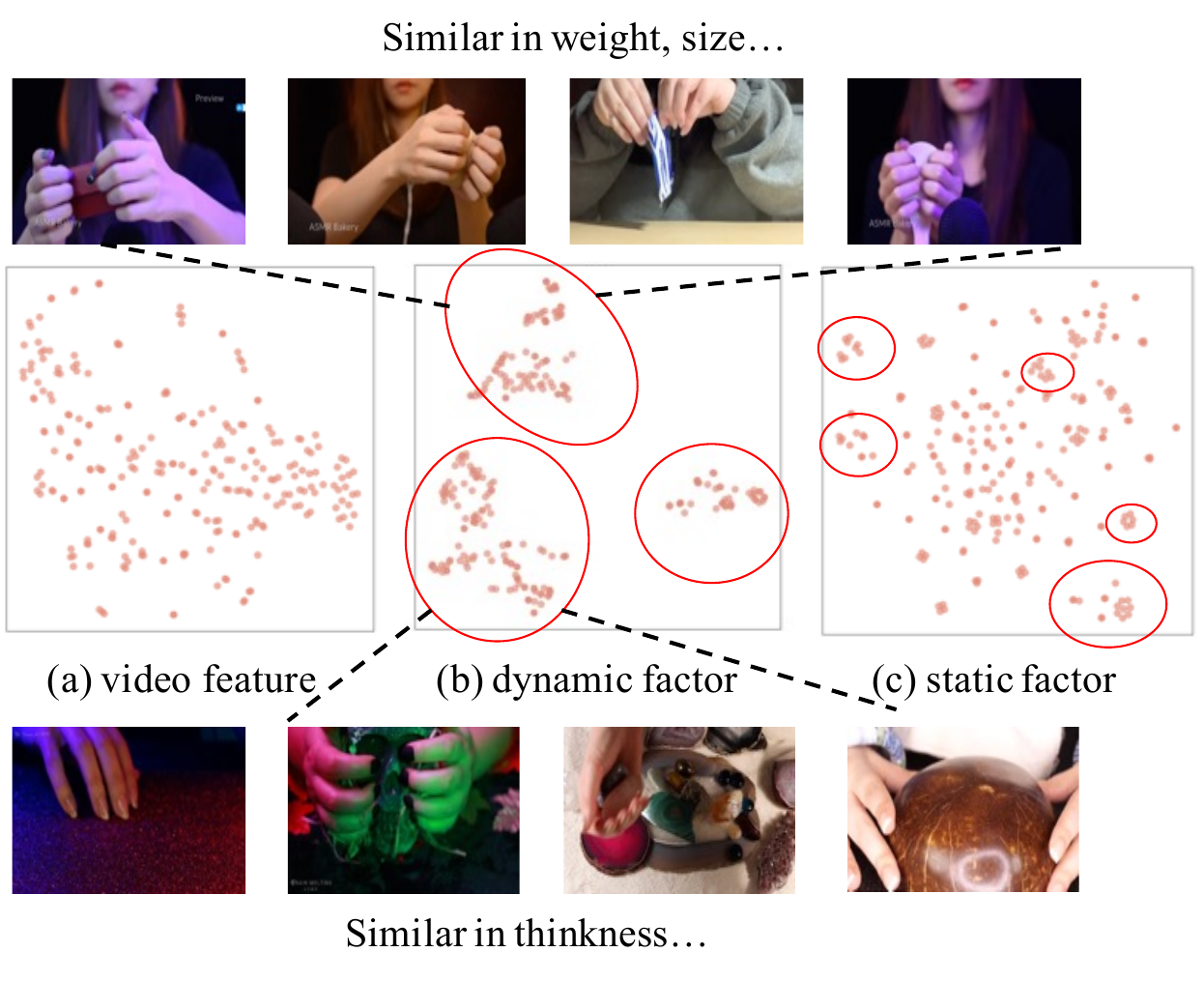}
    \caption{T-SNE visualization of video features before applying DSE+ (a), along with dynamic factors (b) and static factors (c) obtained after DSE+. The \textcolor{red}{red circles} indicate clusters that have been manually identified as containing samples with similar physical properties. We provide examples of these clusters based on shared attributes, including weight and thickness.}
    \label{fig: t-sne}
\end{figure}

Our DSE+ method separates video features into static (time-invariant) and dynamic (time-varying) factors. Figure~\ref{fig: t-sne} shows t-SNE visualizations of these factors alongside raw video features. While raw features appear scattered, dynamic factors extracted by DSE+ exhibit clear clustering, as highlighted by red circles. For example, Figure~\ref{fig: t-sne}(b) shows objects with similar dynamic characteristics, such as small size and lightweight, positioned adjacently. The upper portion of Figure~\ref{fig: t-sne} illustrates a cluster where actions consistently depict a hand grasping and striking the object, reflecting their lightweight characteristics. In contrast to raw features, DSE+ successfully captures this dynamic information. Similarly, the lower section highlights another cluster with shared thickness-related properties, further demonstrating DSE+’s ability to extract dynamic physical characteristics.  
\subsection{Additional Qualitative Results}
\begin{figure*}
    \centering
    \includegraphics[width=0.9\linewidth]{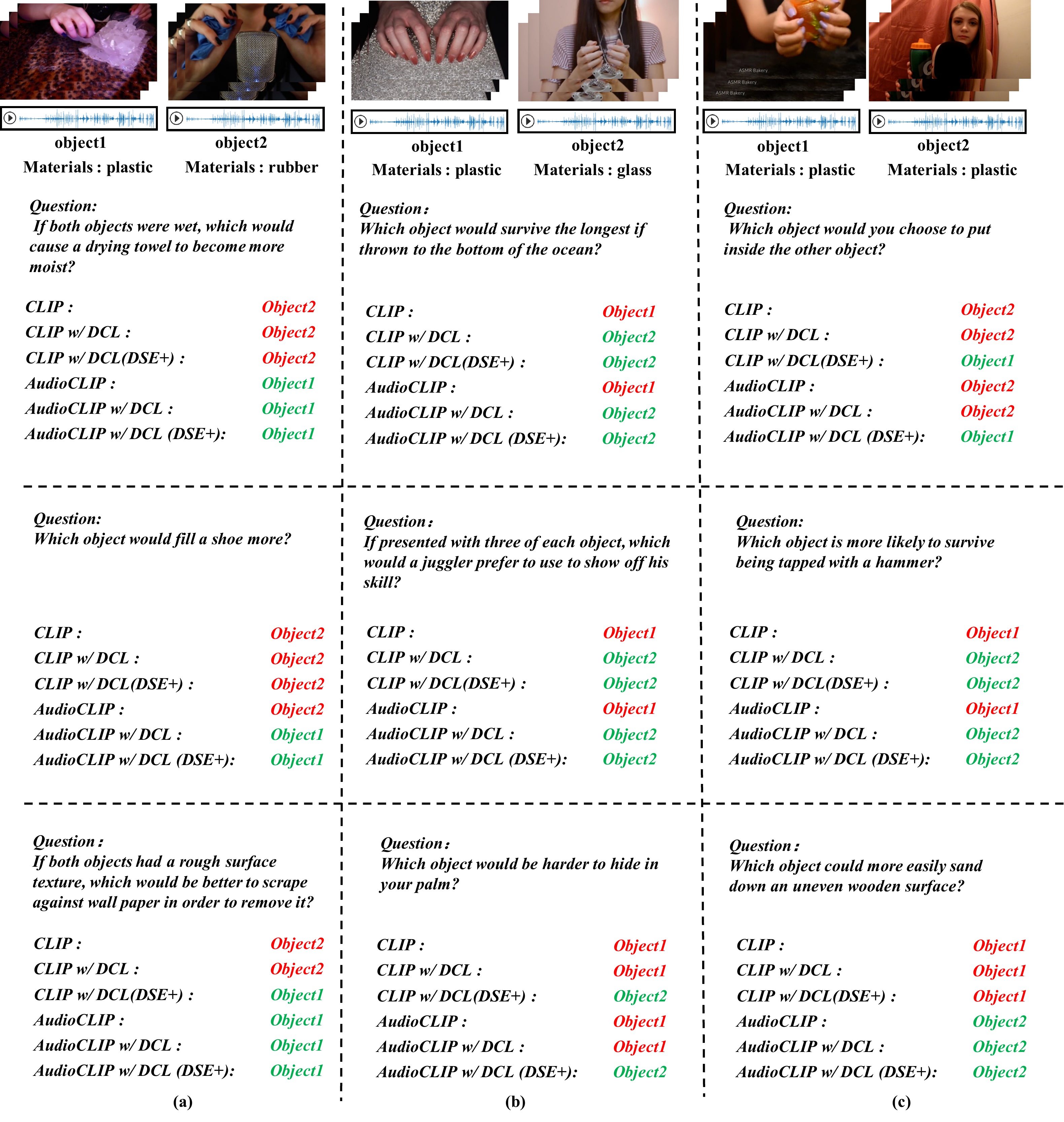}
    \caption{Qualitative Results of baseline w/ and w/o our proposed method, where “Material” refers to the material of the object. The correct answers are depicted in green while the incorrect ones are depicted in red.}
    \label{fig: Qualitative Results_1}
\end{figure*}

As shown in Figure~\ref{fig: Qualitative Results_1}, we present more visualization results comparing our proposed method with other baseline models. It can be seen from the figures that our proposed DCL method outperforms the original process.
\subsection{Impact of visual bias}
\begin{figure}[ht]
    \centering
    \includegraphics[width=0.9\linewidth]{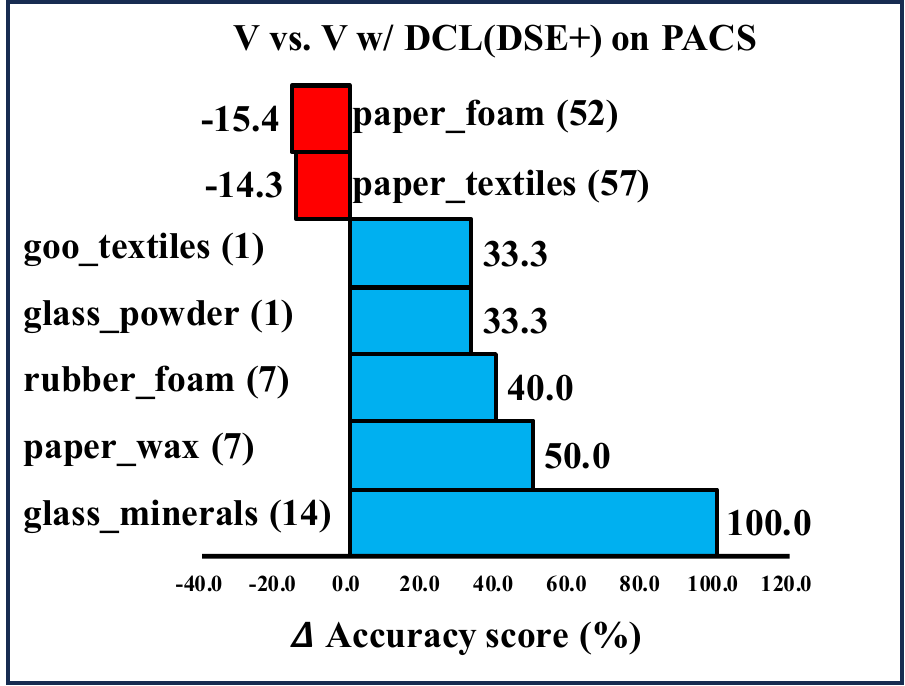}
    \caption{Absolute differences in accuracy scores between two configurations: AudioCLIP with DCL using solely video input (V w/ DCL) and AudioCLIP utilizing only video input (V). The parenthetical value indicates the frequency of occurrence, measured at 11.5 instances within the final 25\% of the training dataset.}
    \label{fig: visual bias}
\end{figure}

As illustrated in Figure~\ref{fig: visual bias}, we show the absolute accuracy differences for specific object pairs. Accuracy for pairs in the lowest 25\% of occurrence frequency improves notably after applying DCL, demonstrating its effectiveness in reducing visual bias for less frequent pairs. However, for some high-frequency pairs ({\it e.g.}, ``paper-foam'' and ``paper-textiles''), a slight accuracy decrease occurs after DCL. This is because dominant visual bias previously led to correct but unreliable predictions, while DCL mitigates this bias, revealing the model’s robust performance.  

\section{VLM-Assisted Reasoning Dataset}
\label{Sec: VLM-Assisted Reasoning Dataset}
\subsection{More examples of VLM-Assisted Reasoning Dataset}
\begin{figure*}
    \centering
    \includegraphics[width=0.75\linewidth]{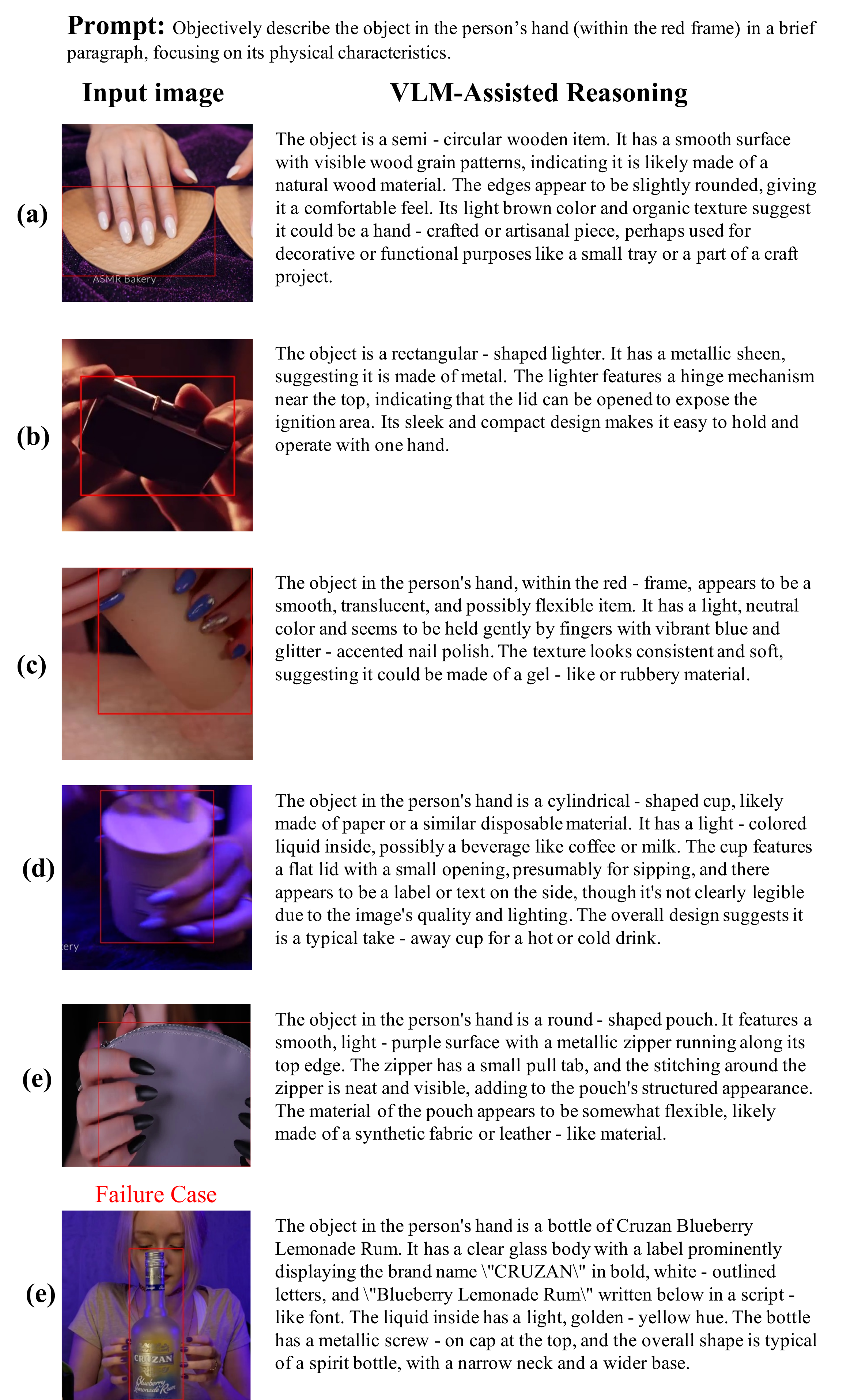}
    \caption{Prompt Text, Input Image, and Corresponding Response of the VLM ({\it i.e.}, Doubao-1.5-Vision-Pro)}
    \label{fig: VLM}
\end{figure*}

Figure \ref{fig: VLM} illustrates the prompts used for the Vision-Language Model (VLM), with subfigures (a)–(f) showcasing its assisted reasoning outputs across various samples. A failure case is evident in Figure~\ref{fig: VLM}(f), where the VLM excessively emphasizes object-specific details ({\it e.g.}, identifying the type of wine) while overlooking the physical characteristics of the glass bottle. Future work will focus on developing more targeted prompting strategies to address such limitations.
\subsection{Dataset Statistics}
\begin{figure}[b]
    \centering
    \includegraphics[width=1.0\linewidth]{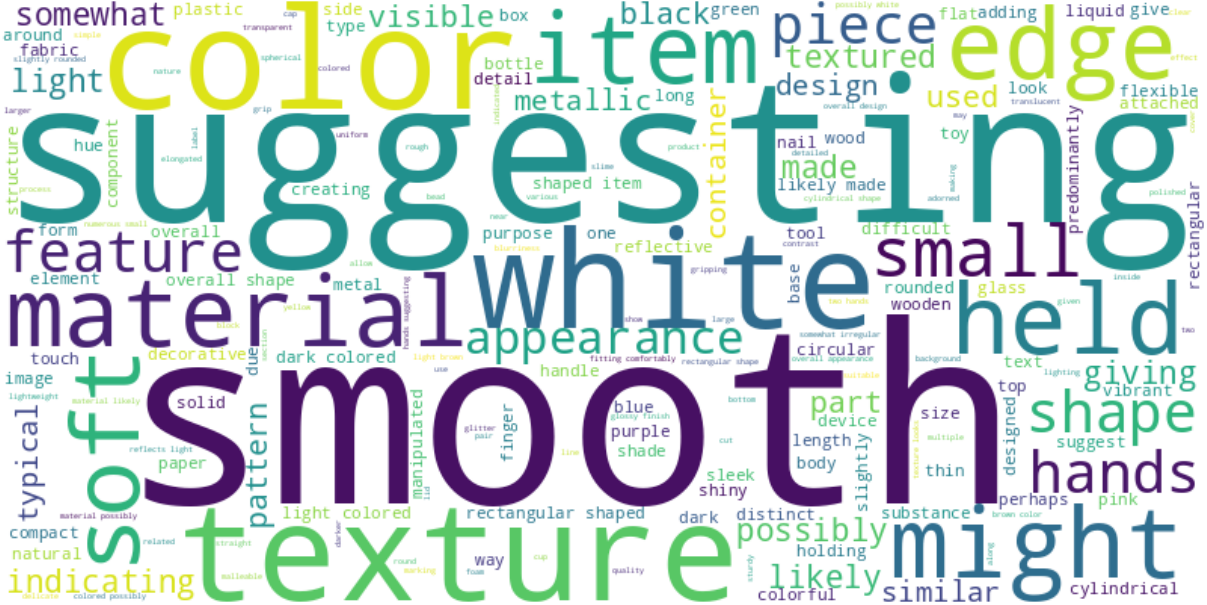}
   \caption{Word Cloud for VLM-Assisted Reasoning}
    \label{fig: Word Cloud}
\end{figure}

\begin{figure}[ht]
    \centering
    \includegraphics[width=1.0\linewidth]{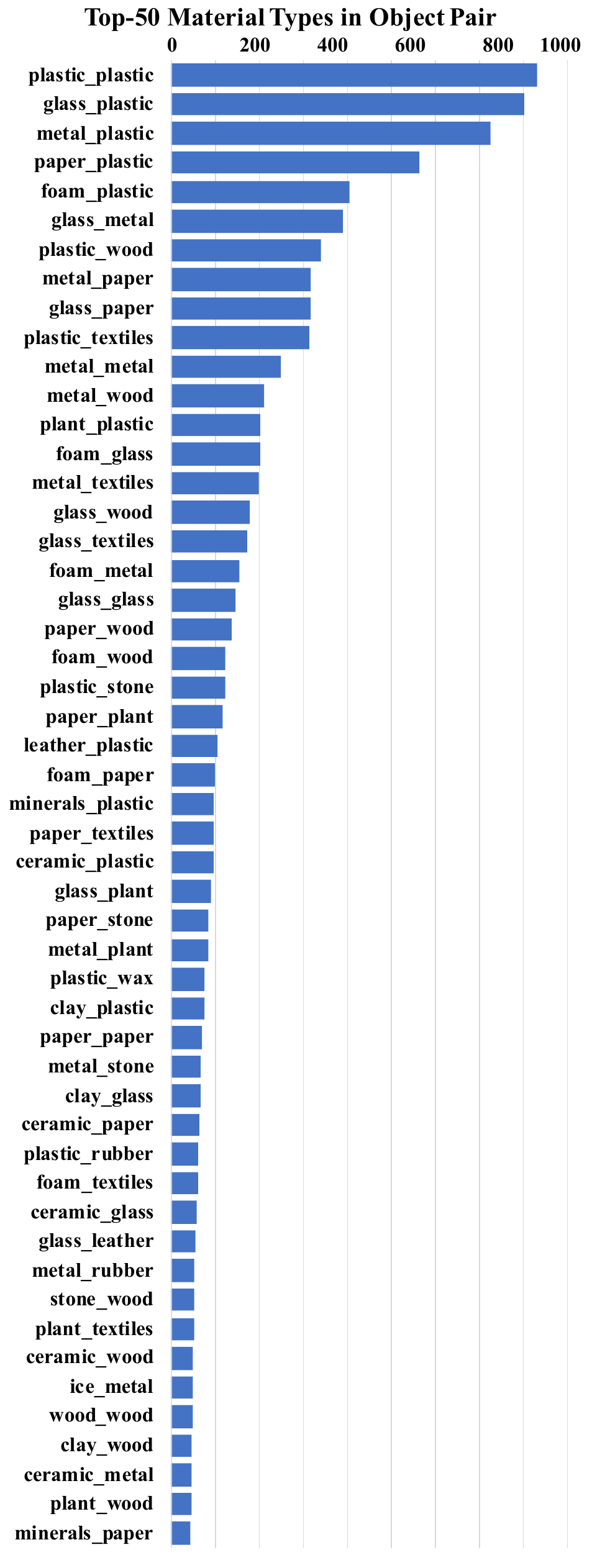}
    \caption{Frequency of Material Types for Object Pairs in the PACS Training Set.}
    \label{fig: Frequency of Material Types}
\end{figure}

We obtained corresponding VLM descriptions for each object in the PACS dataset, resulting in 1,526 descriptions. The average length of these descriptions is 74.05 words, with a maximum length of 118 words and a minimum length of 41 words. The corresponding word cloud is illustrated in Figure~\ref{fig: Word Cloud} and the Top-50 Material Types in the Object Pair are shown in Figure~\ref{fig: Frequency of Material Types}. The generated data is available at https://github.com/MICLAB-BUPT/DCL.
\begin{algorithm}[ht]
\caption{Robust Disentangled Counterfactual Learning (RDCL) Batch-Wise Training}
\label{algorithm: RDCL}
\SetAlgoLined
\KwIn{
    Training batch $\{ \langle v_1, v_2 \rangle_i, \langle a_1, a_2 \rangle_i, q_i \}_{i=1}^{B}$, \\
    Batch size $B$, \\
    Pretrained image encoder $\mathcal{E}_{\text{img}}(\theta)$, \\
    Pretrained audio encoder $\mathcal{E}_{\text{aud}}(\theta)$, \\
    Pretrained text encoder $\mathcal{E}_{\text{text}}(\theta)$, \\
    Labels $\{Y_{GT,i}\}_{i=1}^B$, \\
    Number of frames $T$, proportion of missing data in object 1's video $a_{v1}.$ 
}
\KwOut{
    Predicted answers $\{\hat{Y}_{TIE,i}\}_{i=1}^B$
}

\textbf{Encode features:} \\
\For{$j \in \{1, 2\}$}{ 
   \quad $X^{v_j} = \{X_1^{v_j}, X_2^{v_j}, \cdots, X_T^{v_j}\} \gets \mathcal{E}_{\text{img}}(v_j)$ \\ 
   \quad $X^{a_j} \gets \mathcal{E}_{\text{aud}}(a_j)$
}
$X^t \gets \mathcal{E}_{\text{text}}(q)$ \\

Obtain the set of missing data set $B_{{miss}}$ and the complete data set $B_{{com}}$ using Eq.(27). \\

For each sample $i$ in the batch:

\If{$i \in B_{com}$}{
    \For{each sample in the $B_{{com}}$}{
        Disentangle static factors $X^v_s$ and dynamic factors $X^v_z$ from $X^v$ via DSE in Section 4.2. \\
    }
    
    Use unique encoder and shared encoder to encode $X^v_s$, $X^v_z$, and $X^a$ using Eqs.(24) and (25), obtaining $r_m^{unique}$ and $r_m^{share}, m \in \{a,z,s\}$.
}
\Else{
    Use Eqs. (29) and (30) to complete the missing information.
}

Compute the adjacency matrix $A_X$ using Eqs. (15), (16), and (17). \\

Obtain the fused features $F_1$ and $F_2$ using Eq.(14). \\

Construct the intervened features $X^*$ using Eq. (20), and compute the intervened adjacency matrix ${A}^*$ using Eqs. (15), (16), and (17). \\

Predict $\hat{Y}_{X, A_X}$ and $\hat{Y}_{X^*, A_{X^*}}$ using Eq.(18). \\

Use $\hat{Y}_{TIE}$ obtained from Eq.(19) as the output.
\end{algorithm}

\begin{figure*}
    \centering
    \includegraphics[width=0.9\linewidth]{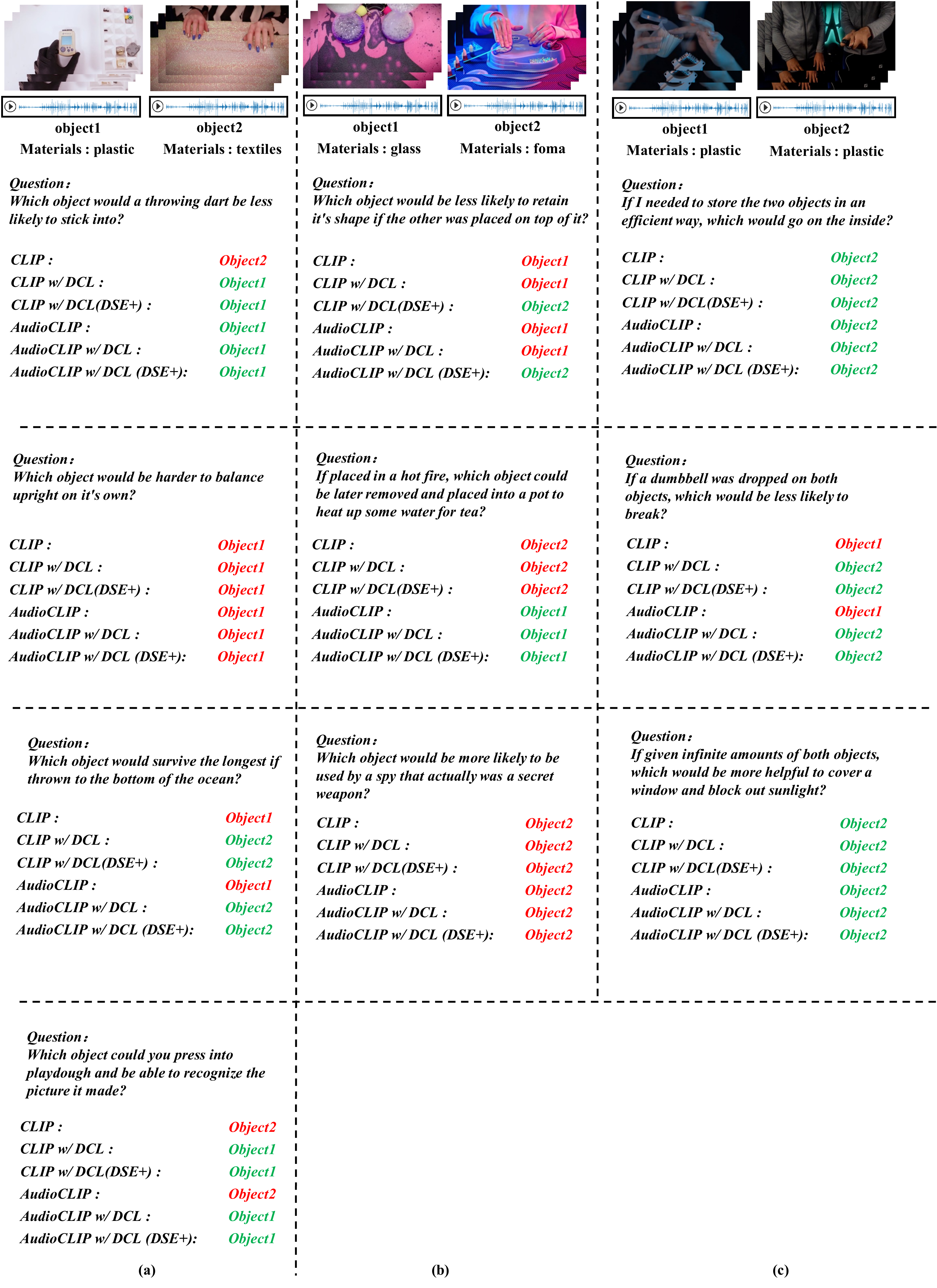}
    \caption{Qualitative Results of baseline w/ and w/o our proposed method, where “Material” refers to the material of the object. The correct answers are depicted in green while the incorrect ones are depicted in red.}
    \label{fig: Qualitative Results_2}
\end{figure*}

\ifCLASSOPTIONcaptionsoff
  \newpage
\fi

\bibliographystyle{IEEEtran}
\bibliography{reference}